\ifdefined\pdfminorversion
\fi
\ifdefined\pdfobjcompresslevel
\fi

\documentclass[10pt,journal,compsoc]{IEEEtran}

\usepackage[utf8]{inputenc}
\usepackage{amsmath,amssymb,amsthm}
\usepackage{graphicx}
\usepackage{booktabs}
\usepackage{multirow}
\usepackage{makecell}
\usepackage{array}
\usepackage{float}
\usepackage{pifont}
\usepackage{ragged2e}
\usepackage{url}
\usepackage{xcolor}
\usepackage{colortbl}
\usepackage{algorithm}
\usepackage{algorithmic}

\ifCLASSOPTIONcompsoc
  \usepackage[nocompress]{cite}
\else
  \usepackage{cite}
\fi

\usepackage[colorlinks=true,linkcolor=blue,citecolor=blue,urlcolor=blue]{hyperref}

\definecolor{headergray}{RGB}{245,245,245}
\definecolor{rowblue}{RGB}{255,244,235}
\definecolor{colavg}{RGB}{248,248,250}
\definecolor{rowgreen}{RGB}{240,248,243}
\definecolor{textpurple}{RGB}{255,0,0}

\newcommand{\erahighlight}[1]{#1}
\newcommand{\pub}[1]{$_{\text{#1}}$}
\newcommand{\cmark}{\ding{51}}

\graphicspath{{images/}}

\begin{document}

\title{ERA: Entropy-Guided Visual Token Pruning with Rectified Attention for Efficient MLLMs}

  \author{Yuhao~Wang,
    Mu~Qiao,
    Haiwen~Diao$^{\dagger,\,\ast}$,
    Yunzhi~Zhuge$^{\dagger}$,
    Pingping~Zhang$^{\dagger}$, ~\IEEEmembership{Member,~IEEE},\\
    Xindong~Zhang,
    Lei~Zhang,~\IEEEmembership{Fellow,~IEEE} and
    Huchuan~Lu,~\IEEEmembership{Fellow,~IEEE}
\IEEEcompsocitemizethanks{
\IEEEcompsocthanksitem $\dagger$ denotes the corresponding author. $\ast$ denotes the project lead.
\IEEEcompsocthanksitem Yuhao Wang, Mu Qiao, Yunzhi Zhuge, Pingping Zhang, and Huchuan Lu are with the School of Future Technology, Dalian University of Technology, Dalian 116024, China (e-mail: 924973292@mail.dlut.edu.cn; q15841685383@mail.dlut.edu.cn; zgyz@dlut.edu.cn; zhpp@dlut.edu.cn; lhchuan@dlut.edu.cn).
\IEEEcompsocthanksitem Haiwen Diao is with S-Lab, Nanyang Technological University, Singapore 639798 (e-mail: haiwen.diao@ntu.edu.sg).
\IEEEcompsocthanksitem Xindong Zhang is with the OPPO Research Institute, Shenzhen 518000, China (e-mail: zhangxindong1@oppo.com).
\IEEEcompsocthanksitem Lei Zhang is with the Department of Computing, The Hong Kong Polytechnic University, Hong Kong (e-mail: cslzhang@comp.polyu.edu.hk).
}}

\markboth{IEEE Transactions on Pattern Analysis and Machine Intelligence}%
{Wang \MakeLowercase{\textit{et al.}}: ERA: Entropy-Guided Visual Token Pruning with Rectified Attention for Efficient MLLMs}

\IEEEtitleabstractindextext{%
\begin{abstract}
\justifying
Multimodal Large Language Models (MLLMs) incur prohibitive inference costs due to long visual token sequences.
Training-free visual token reduction provides an efficient solution.
However, existing methods distort attention distributions, giving rise to a phenomenon we term \emph{Attention Logit Collapse}.
To address this issue, we propose \textbf{ERA}, an \textbf{E}ntropy-guided visual token pruning framework with \textbf{R}ectified \textbf{A}ttention for efficient MLLMs.
Specifically, ERA comprises three crucial components: \textbf{Dual-view Entropy Pruning (DEP)}, \textbf{Bias-aware Token Recycling (BTR)}, and \textbf{Logit-preserving Attention Rectification (LAR)}.
First, DEP identifies representative anchor tokens by jointly modeling visual diversity and head-wise saliency.
BTR then recycles pruned tokens into their corresponding anchors while estimating a cluster-level logit bias.
Building upon this, LAR injects the estimated bias into attention logits, effectively rectifying the collapse induced by token reduction.
Together, these components preserve visual evidence even under aggressive compression, enabling robust performance across single-image, multi-image, and video settings on a wide range of MLLMs.
Beyond delivering practical acceleration, ERA establishes logit-preserving visual token pruning as a principled framework for efficient MLLMs, unifying theoretical foundation, algorithmic design, and practical deployment.
The code is at \url{https://github.com/924973292/ERA}.
\end{abstract}

}
\maketitle
\IEEEdisplaynontitleabstractindextext
\IEEEpeerreviewmaketitle

\section{Introduction}
\label{sec:intro}
\IEEEPARstart{M}{ultimodal} Large Language Models (MLLMs) extend Large Language Models (LLMs) to the visual domain~\cite{liu2024improved} by integrating visual tokens into the language backbone, achieving strong performance on diverse tasks.
However, modern vision encoders~\cite{chen2024internvl} generally include hundreds to thousands of visual tokens, making computation and memory scale poorly~\cite{chen2024image}.
This redundancy becomes more severe for high-resolution images, multi-image inputs and video sequences, where visual tokens dominate the prefill stage and inflate the cache.
Thus, training-free visual token reduction~\cite{shao2025tokens} has become a practical solution for accelerating MLLM inference without retraining the model.

Recently, great progress has been made in this direction.
For example, FastV~\cite{chen2024image} shows that many visual tokens can be removed after early LLM layers according to attention statistics.
DivPrune~\cite{alvar2025divprune} and CDPruner~\cite{zhang2025beyond} improve token selection by emphasizing diversity and instruction relevance.
VisionZip~\cite{yang2025visionzip} further combines token selection with token merging to preserve compact visual representations.
These methods demonstrate that substantial visual token redundancy exists in MLLM inference.

\begin{figure*}[t]
  \centering
  \includegraphics[width=1\linewidth]{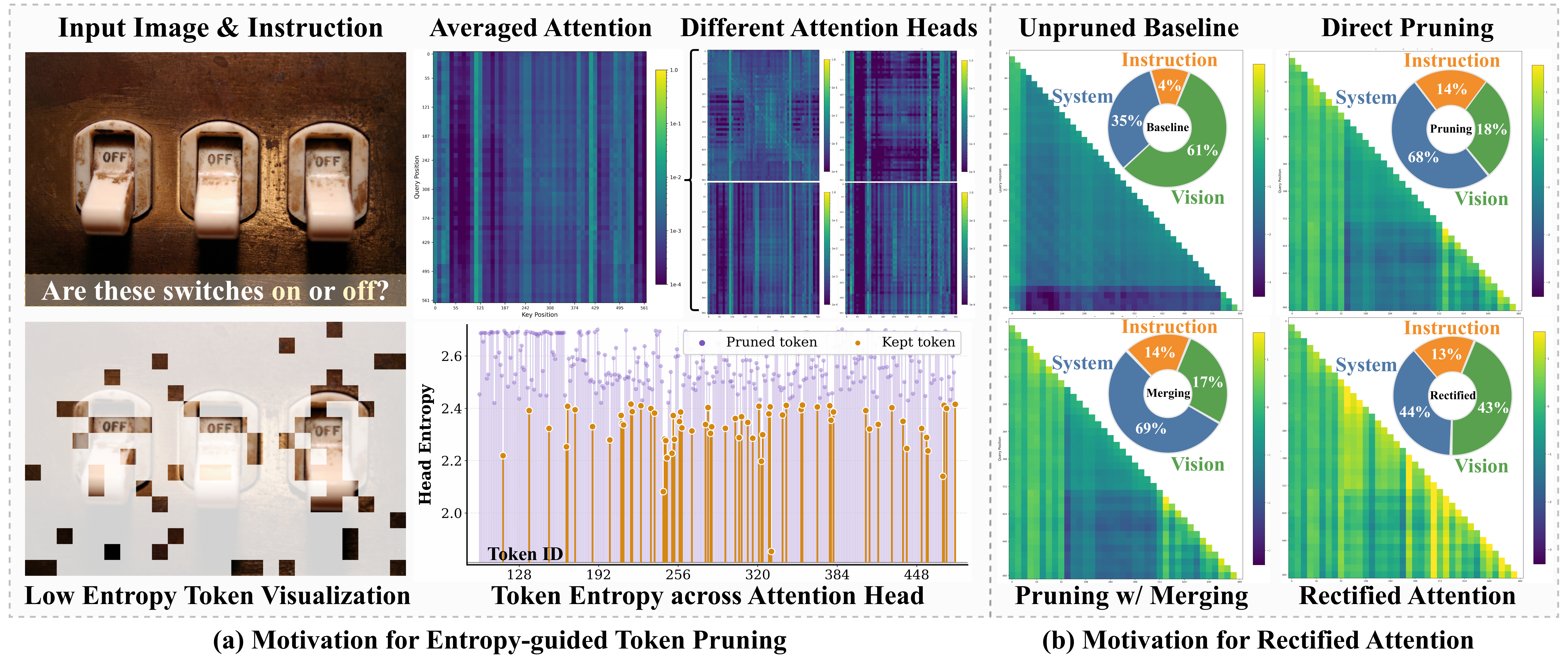}
    \vspace{-7mm}
\caption{Motivations for ERA with LLaVA-1.5-7B on the VQA$^{\text{T}}$ dataset.
\textbf{(a) Token entropy reveals head-wise saliency.} Averaged attention can obscure tokens that are critical to specific heads. Head-wise entropy exposes these low-entropy, head-specific responses concentrated on salient regions, providing an auxiliary criterion for importance estimation.
\textbf{(b) Attention logit collapse under token reduction.} Across the four settings, pruning and merging reduce the fraction of attention logit assigned to visual tokens and shift attention toward system and instruction tokens. Our proposed rectified attention realigns this distribution with the full-sequence baseline, preserving visual evidence after compression.}
  \label{fig:motivation}
    \vspace{-3mm}
\end{figure*}

Despite great progress, existing methods suffer from two fundamental limitations.
The first lies in token importance estimation.
Most methods prune or merge tokens using a single criterion, such as visual diversity~\cite{alvar2025divprune} or sparsity~\cite{chen2024image}.
Even when the diversity and sparsity are combined~\cite{yang2025visionzip}, these methods overlook the attention patterns across different heads~\cite{chowdhury2025prompt}.
As shown in \erahighlight{Fig.~\ref{fig:motivation} (a)}, the averaged attention can obscure tokens that are critical to specific heads.
We find that such head-specific tokens exhibit low entropies in their attention distributions.
They concentrate on salient regions, motivating information entropy as an auxiliary criterion for token importance estimation.
More importantly, we identify a previously overlooked phenomenon in current MLLMs, termed \emph{Attention Logit Collapse}.
In \erahighlight{Fig.~\ref{fig:motivation} (b)}, we compare the distribution of attention logits across token types under four settings.
In the baseline with a full sequence, visual tokens account for 61\% of the attention logit, whereas system and instruction tokens account for 35\% and 4\%, respectively.
After pruning, the distribution shifts significantly.
The fraction of attention logits assigned to visual tokens drops to 18\%, whereas system and instruction tokens increase to 68\% and 14\%, respectively.
This shift suggests that pruning suppresses visual attention and highlights non-visual tokens.
Crucially, this distortion persists even with token merging.
Theoretically, this collapse is attributed to how the attention logit is aggregated.
In practice, the attention logit for a given region equals the sum of exponentiated logits over all tokens in that region.
However, existing token reduction methods approximate this sum with a single centroid logit, thereby systematically underestimating the accumulated attention logit of visual tokens and leading to the collapse.

Motivated by these observations, we propose \textbf{ERA}, an \emph{\textbf{E}ntropy-guided visual token pruning framework with \textbf{R}ectified \textbf{A}ttention} for efficient MLLM inference.
Specifically, ERA comprises three key components: \textbf{Dual-view Entropy Pruning (DEP)}, \textbf{Bias-aware Token Recycling (BTR)}, and \textbf{Logit-preserving Attention Rectification (LAR)}.
First, DEP identifies representative anchor tokens in a unified space that jointly models visual diversity and head-wise saliency, enabling the retained tokens to cover both complementary visual patterns and head-specific evidence.
Second, BTR recycles pruned tokens into their corresponding anchors and estimates a cluster-level logit bias, preserving discarded visual evidence while recording its accumulated contribution to attention.
Third, LAR injects the estimated bias into attention logits, rectifying the collapse induced by token reduction.
This keeps the attention computed on reduced tokens consistent with the original distribution.
Crucially, LAR admits a kernel-friendly realization via matrix augmentation and remains compatible with optimized attention implementations such as FlashAttention~\cite{dao2022flashattention}.
To validate the effectiveness of ERA, we conduct extensive experiments across single-image, multi-image and video settings, covering representative MLLMs with different visual-token generation paradigms.
Beyond standard benchmark evaluation, we further assess ERA in a practical vLLM~\cite{kwon2023efficient} serving environment to verify the transferability of its efficiency gains to real-world deployment.
Across these settings, ERA remains robustness under aggressive token reduction while delivering consistent acceleration benefits.

In summary, our contributions are as follows:
\begin{itemize}
    \item We reveal two observations for training-free visual token reduction methods in MLLMs.
    The head-wise attention entropy provides an effective criterion for identifying tokens that are salient to specific attention heads. 
    More importantly, we identify a previously overlooked phenomenon, termed \emph{Attention Logit Collapse}, in which existing methods systematically underestimate the accumulated attention logit of visual tokens through cluster-based approximation.
    \item We propose ERA, an entropy-guided visual token pruning framework with rectified attention for efficient MLLM inference.
    It first uses Dual-view Entropy Pruning (DEP) to select anchors with both diversity and saliency.
    Then, Bias-aware Token Recycling (BTR) recycles pruned tokens while estimating a cluster-level logit bias.
    Finally, Logit-preserving Attention Rectification (LAR) injects the estimated bias to restore the compressed attention distribution.
    \item We conduct extensive experiments across single-image, multi-image and video settings on representative MLLMs with different visual-token generation paradigms.
    The results show that ERA remains robustness under aggressive token reduction while delivering inference gains in FLOPs, latency, prefill speed and cache usage, including deployment-oriented improvements with vLLM.
\end{itemize}

\section{Related Work}
\label{sec:related_work}
\subsection{Multimodal Large Language Models}
MLLMs enable unified perception and reasoning by connecting visual representations with LLMs through diverse architectural designs~\cite{baltrusaitis2019multimodal,xu2023multimodal,zhang2024visionlanguage}.
Given that the attention computation scales quadratically with sequence length, how visual tokens are represented and injected into LLMs largely dictates inference efficiency.
Accordingly, current MLLMs can be organized into three categories: \textit{static projection pipelines}, \textit{resolution-adaptive strategies} and \textit{unified architectures}.
\\
\textbf{Static Projection Pipelines.}
This paradigm bridges a vision encoder with an LLM via a learnable projector, mapping patch-level features into the LLM semantic space.
LLaVA~\cite{liu2024improved} and MiniGPT-4~\cite{zhu2024minigpt} establish this baseline by aligning visual features with language tokens.
InternVL~\cite{chen2024internvl} scales this approach to foundation models with stronger encoders.
Despite their effectiveness, these methods typically employ a static projection for visual tokens, resulting in prohibitive computation costs and memory usage, especially for high-resolution scenarios~\cite{chen2024image,chen2024longvila} and video inputs~\cite{li2024llavaonevision}.
\\
\textbf{Resolution-Adaptive Strategies.}
To mitigate information loss and redundancy caused by fixed-size resizing, some approaches adapt visual tokenization to image resolution and aspect ratio.
For example, LLaVA-NeXT~\cite{liu2024llavanext} and InternLM-XComposer2~\cite{dong2024internlm} utilize dynamic grid pinning to enhance fine-grained perception capabilities.
Qwen2.5-VL~\cite{bai2025qwen2} studies dynamic-resolution tokenization to support any-resolution perception.
While improving visual fidelity, these methods may sharply increase visual tokens with high-resolution inputs, amplifying the need for visual token reduction under constrained computational budgets.
\\
\textbf{Unified Architectures.}
Different from the modular designs, unified architectures integrate vision and language processing within a single backbone.
For example, EVE series~\cite{diao2024unveiling,diao2025evev2} and the NEO family~\cite{diao2025pixels,diao2026pixels,diao2026sensenova} process visual and textual tokens as a single sequence, requiring no explicit projectors or vision encoders.
However, they typically process visual tokens without intrinsic token reduction.
Thus, substantial computational redundancy remains, underscoring the need for more advanced visual token reduction methods to improve efficiency in unified MLLMs.
\subsection{Visual Token Reduction for MLLMs}
Many visual tokens introduce substantial computational and memory overhead in MLLMs~\cite{han2023survey,papa2024efficient}.
Recently, visual token reduction draws attention for accelerating MLLM inference.
Specifically, existing visual token reduction methods can be categorized into three paradigms: \textit{attention-based pruning}, \textit{similarity-based pruning} and \textit{merging-based pruning}.
\\
\textbf{Attention-based Pruning.}
Attention-based pruning estimates token importance from attention distributions or cross-modal interactions.
For example, FastV~\cite{chen2024image} prunes visual tokens of early LLM layers based on attention statistics.
SparseVLM~\cite{zhang2024sparsevlm} leverages cross-modal attention to preserve instruction-relevant visual evidence.
Besides, VTW~\cite{lin2025boosting} exploits the attention sink phenomenon to withdraw visual tokens in deeper layers of LLMs. 
FitPrune~\cite{ye2025fit} improves the pruning stability by minimizing the divergence of attention distributions.
Although effective, attention-based methods are often context-dependent and may not be fully compatible with lightweight attention kernels such as FlashAttention~\cite{dao2022flashattention}, thereby limiting the inference efficiency of MLLMs.
\\
\textbf{Similarity-based Pruning.}
Similarity-based pruning reduces redundancy by selecting a compact subset of visual tokens according to feature diversity.
For example, DART~\cite{wen2025stop} ranks and selects tokens using diversity-aware scores to retain representative visual content.
DivPrune~\cite{alvar2025divprune} emphasizes feature dispersion to reduce redundancy among visual tokens.
However, similarity-based pruning may ignore instruction relevance.
CDPruner~\cite{zhang2025beyond} addresses this issue by selecting tokens that are both diverse and relevant to the instruction with a subset selection~\cite{kulesza2012determinantal}.
Nevertheless, both attention-based and similarity-based pruning discard redundant tokens, resulting in irreversible information loss and motivating token merging as an alternative paradigm.
\\
\textbf{Merging-based Pruning.}
Merging-based pruning reduces redundancy by aggregating or resampling visual tokens into compact representations within MLLMs.
For example, BLIP-2~\cite{li2023blip} uses a Q-Former to distill visual features into a fixed number of query tokens.
VisionZip~\cite{yang2025visionzip} merges similar visual tokens to extend the effective context length.
LLaVA-PruMerge~\cite{shang2025llava} combines pruning and merging to dynamically adapt the visual token count at inference time.
While effective, merging-based pruning methods commonly overlook the phenomenon of \emph{Attention Logit Collapse}, leading to information loss during token reduction.
In contrast, our proposed framework explicitly preserves attention logits via entropy-guided pruning and attention rectification, achieving superior trade-offs between efficiency and accuracy.
\section{Method}
\label{sec:method}
\begin{figure*}[t]
  \centering
  \includegraphics[width=1\linewidth]{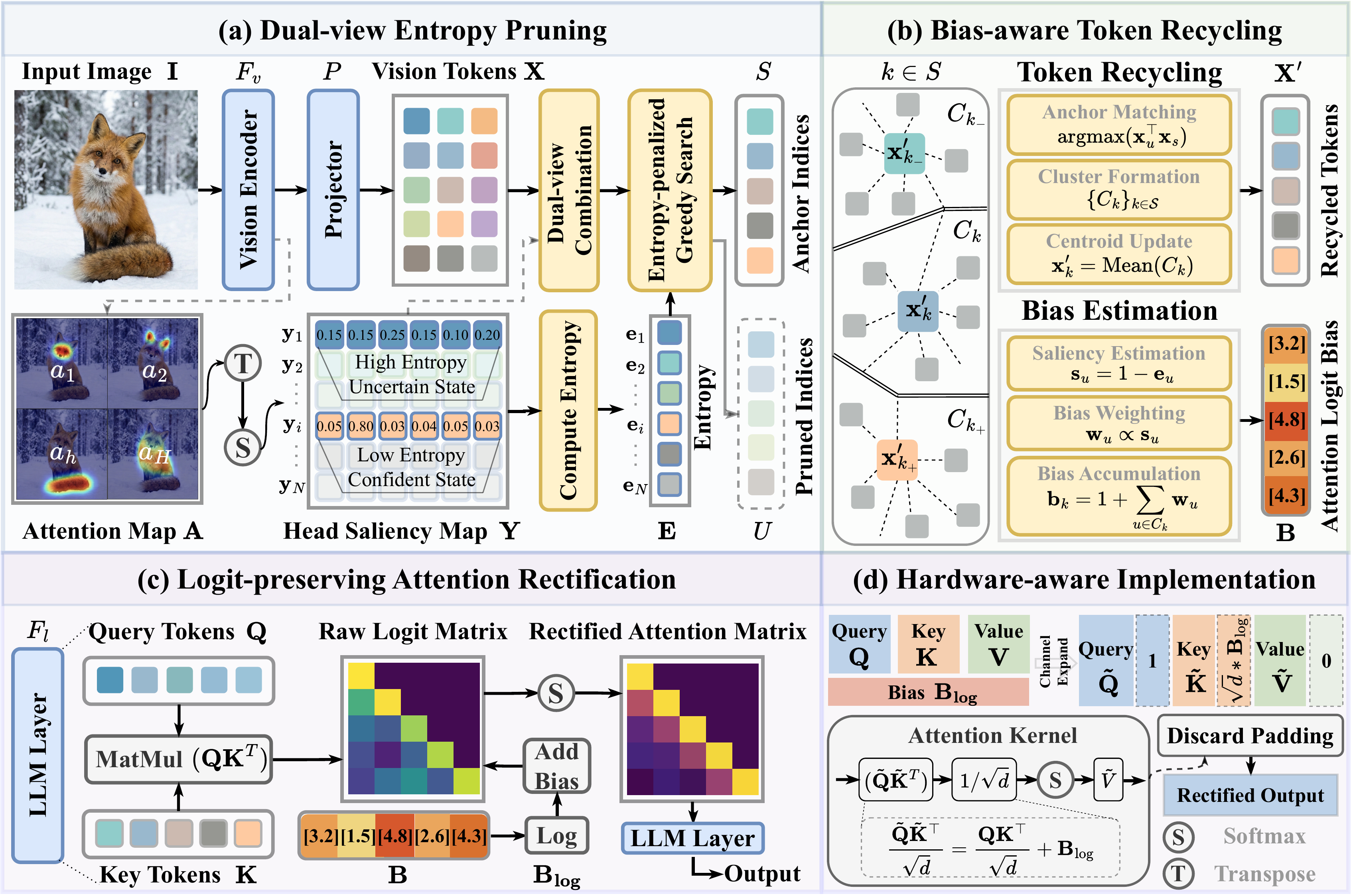}
  \vspace{-2mm}
  \caption{
  Overview of the proposed ERA framework. 
  ERA consists of three synergistic components.
  (a) \textbf{Dual-view Entropy Pruning (DEP)} selects anchors by jointly modeling visual diversity and head-wise saliency. 
  (b) \textbf{Bias-aware Token Recycling (BTR)} merges pruned tokens into their nearest anchors and estimates a cluster-level logit bias. 
  (c) \textbf{Logit-preserving Attention Rectification (LAR)} injects the estimated bias to rectify the \textit{Attention Logit Collapse}. 
  (d) \textbf{Hardware-aware Implementation} leverages matrix augmentation to maintain compatibility with optimized attention kernels. 
  With these components, ERA enables aggressive token reduction while preserving robust and efficient MLLM inference.
  }
\label{fig:overall}
\vspace{-2mm}
\end{figure*}

We propose ERA, an entropy-guided token pruning framework with rectified attention for efficient MLLM inference.
As illustrated in \erahighlight{Fig.~\ref{fig:overall}}, ERA comprises three synergistic components:
\noindent(1) \textbf{Dual-view Entropy Pruning} (\erahighlight{\S\ref{ssec:sampling}}) identifies a robust anchor set by balancing visual diversity with saliency derived from head-wise attention entropy.
\noindent(2) \textbf{Bias-aware Token Recycling} (\erahighlight{\S\ref{ssec:recycling}}) aggregates pruned tokens into their nearest anchors to form recycled representations and estimates an effective bias for each cluster.
\noindent(3) \textbf{Logit-preserving Attention Rectification} (\erahighlight{\S\ref{ssec:rectification}}) injects the estimated bias into attention logits, ensuring that the LLM's interaction with the reduced tokens remains consistent with the original distribution.
Details are described as follows.
\subsection{Dual-view Entropy Pruning}
\label{ssec:sampling}
Effective token pruning benefits from anchor tokens that can preserve visual diversity and capture attentive information.
Relying solely on visual diversity may include irrelevant background~\cite{zhang2025beyond}, while relying solely on attention may result in redundancy~\cite{tong2025flowcut}.
To address this issue, we introduce Dual-view Entropy Pruning (DEP), which operates in a unified space that couples visual appearance with head-wise saliency for robust anchor selection.
As shown in \erahighlight{Fig.~\ref{fig:overall} (a)}, given an input image $\mathbf{I}$, the vision encoder $F_v$ produces visual tokens $\mathbf{X} \in \mathbb{R}^{N \times D}$.
Here, $N$ is the number of visual tokens and $D$ is the embedding dimension.
Meanwhile, we extract the attention map $\mathbf{A} \in \mathbb{R}^{H \times N}$ from the vision encoder's CLS token over $H$ heads.
Then, we define the head saliency map $\mathbf{Y} \in \mathbb{R}^{N \times H}$ for each token via a softmax operation with the following equation:
\begin{equation}
    \mathbf{y}_i = \text{Softmax}(\mathbf{A}^\top[i, :]).
\end{equation}
Here, $\mathbf{y}_i \in \mathbb{R}^{H}$ captures how token $i$ distributes its importance across different attention heads.
We then compute the entropy $\mathbf{E}$ of this distribution to quantify the token saliency:
\begin{equation}
    \mathbf{e}_i = -\sum_{h=1}^{H} \mathbf{y}_{i,h} \log (\mathbf{y}_{i,h} + \epsilon).
\end{equation}
Here, $\epsilon$ is a small constant for numerical stability.
Next, we perform a dual-view combination by concatenating the visual token $\mathbf{x}_i$ with the saliency feature $\mathbf{y}_i$, followed by $\ell_2$ normalization to obtain $\mathbf{z}_{i} \in \mathbb{R}^{D+H}$ as follows:
\begin{equation}
    \mathbf{z}_i =
    \frac{\text{Concat}\left( \mathbf{x}_i, \alpha \mathbf{y}_i \right)}
    {\left\| \text{Concat}\left( \mathbf{x}_i, \alpha \mathbf{y}_i \right) \right\|_2},
\end{equation}
where $\alpha$ balances these two views.
Using this representation, we adopt an entropy-penalized greedy search to construct an anchor set $S$ that balances diversity and saliency, as illustrated in \erahighlight{Alg.~\ref{alg:sampling}}.
Specifically, we initialize $S$ with the minimum-entropy token as the starting anchor.
At each step, we select the candidate that is farthest from the current anchors in the dual-view space, while penalizing high-entropy tokens with a factor $\eta$ with the following equation:
\begin{equation}
    j^* = \arg\max_{j \notin S}\left(-\max_{i \in S} \mathbf{z}_j^\top \mathbf{z}_i - \eta \mathbf{e}_j\right).
\end{equation}
The similarity term promotes the coverage of complementary visual patterns and saliency distributions, while the entropy penalty suppresses tokens whose head-wise importance is diffuse.
We repeat the above update until reaching the target anchor size $K$.
After selection, we obtain the anchor indices $S$ and define the pruned indices as its complement $U = \{1,\ldots,N\} \setminus S$.
Next, we recycle tokens in $U$ by merging them into their nearest anchors in $S$ to preserve information as described in the following section.
\subsection{Bias-aware Token Recycling}
\label{ssec:recycling}
Anchor selection is insufficient for effective token reduction, as naively discarding pruned tokens inevitably leads to information loss.
To address this issue, we propose Bias-aware Token Recycling (BTR), which recycles pruned tokens into anchors while explicitly estimating the accumulated logit bias of each cluster for subsequent attention rectification.
\noindent\textbf{Anchor Matching and Cluster Formation.}
As shown in \erahighlight{Fig.~\ref{fig:overall} (b)}, given the anchor set $S$ and the pruned set $U$, we assign each pruned token to its nearest anchor based on the similarity of visual features.
Specifically, for each $u \in U$, we identify its matched anchor with the following equation:
\begin{equation}
    \phi(u) = \arg\max_{s \in S} \left( \mathbf{x}_u^\top \mathbf{x}_s \right).
\end{equation}
This induces a set of clusters $\{C_k\}_{k \in S}$, where each cluster contains one anchor and its associated pruned tokens:
\begin{equation}
    C_k = \{k\} \cup \{u \in U \mid \phi(u) = k\}.
\end{equation}
\noindent\textbf{Centroid Update.}
After clustering, we compress each cluster into a single recycled token by updating the anchor representation.
To preserve semantic information, the token recycling is performed in the embedding space as follows:
\begin{equation}
    \mathbf{x}_k^{\prime} = \frac{1}{|C_k|} \sum_{i \in C_k} \mathbf{x}_i .
\end{equation}
This produces a recycled token sequence $\mathbf{X}' = \{\mathbf{x}_k'\}_{k \in S}$ that semantically approximates the full visual token set.
\begin{algorithm}[t]
\caption{Entropy-penalized Greedy Search}
\label{alg:sampling}
\begin{algorithmic}[1]
\STATE \textbf{Input:} Dual-view representation $\mathbf{Z} \in \mathbb{R}^{N \times (D+H)}$, Entropy $\mathbf{E} \in \mathbb{R}^{N}$, Balance factor $\eta$, Target size $K$
\STATE \textbf{Output:} Anchor indices $S$
\STATE Compute similarity matrix $\mathbf{C} \leftarrow \mathbf{Z}\mathbf{Z}^\top$
\STATE Initialize $s_0 \leftarrow \arg\min_i \mathbf{e}_i$; $S \leftarrow \{s_0\}$
\STATE Initialize cache $\mathbf{m} \leftarrow \mathbf{C}[:, s_0]$
\WHILE{$|S| < K$}
    \STATE Select anchor: $j^* \leftarrow \arg\max_{j \notin S}\left(-\mathbf{m}_j - \eta\,\mathbf{e}_j\right)$
    \STATE Update $S \leftarrow S \cup \{j^*\}$
    \STATE Update cache: $\mathbf{m} \leftarrow \max\big(\mathbf{m}, \mathbf{C}[:, j^*]\big)$
\ENDWHILE
\STATE \textbf{return} $S$
\end{algorithmic}
\end{algorithm}
\\
\noindent\textbf{Bias Estimation.}
To estimate the attention logit bias accumulated by each cluster, we first normalize the entropy of each pruned token within the current visual sequence and convert it into a bounded saliency score as follows:
\begin{equation}
    \bar{\mathbf{e}}_u =
    \frac{\mathbf{e}_u - \min_i \mathbf{e}_i}
    {\max_i \mathbf{e}_i - \min_i \mathbf{e}_i + \epsilon},
    \quad
    \mathbf{s}_u = 1 - \bar{\mathbf{e}}_u .
\end{equation}
Here, $\mathbf{s}_u \in [0,1]$ and a lower entropy indicates a higher saliency.
We then convert the saliency into a bias weight for aggregating the contribution of each pruned token with the mapping function as follows:
\begin{equation}
    \mathbf{w}_u = \lambda + (1-\lambda)\mathbf{s}_u,
    \quad \lambda \in [0,1].
    \label{eq:bias_weight}
\end{equation}
Finally, the estimated attention logit bias associated with each recycled token is obtained by accumulating the weights from its assigned tokens with the following equation:
\begin{equation}
    \mathbf{b}_k = 1 + \sum_{u \in C_k \setminus \{k\}} \mathbf{w}_u .
    \label{eq:cluster_bias}
\end{equation}
Here, the constant term accounts for the anchor itself, while the summation indicates the bias accumulated from pruned tokens.
The recycled tokens $\mathbf{X}'$ and their associated bias $\mathbf{B}=\{\mathbf{b}_k\}_{k\in S}$ jointly encode the semantic content and the logit bias.
In the next stage, this information serves as the input to attention rectification, enabling efficient inference while preserving the original attention distribution.
\subsection{Logit-preserving Attention Rectification}
\label{ssec:rectification}
The standard dot-product attention mechanism assigns each token a logit, and the subsequent softmax is determined by the exponentiated logit.
After token merging, a single centroid logit no longer captures the accumulated exponentiated logit of all cluster members, causing a mismatch between the reduced and original attention distribution.
As shown in \erahighlight{Fig.~\ref{fig:motivation} (b)}, we refer to this phenomenon as \emph{Attention Logit Collapse}.
To address this issue, we propose Logit-preserving Attention Rectification (LAR), which injects a log-bias term into the attention logit to rectify the collapse and preserve the original distribution.
As illustrated in \erahighlight{Fig.~\ref{fig:overall} (c)}, the ideal exponentiated logit assigned to the cluster $C_k$ in the original sequence at a given LLM layer $F_l$ can be represented as follows:
\begin{equation}
    P_k^{\text{ideal}}
    = \sum_{i \in C_k} \exp\left( \frac{\mathbf{q}^\top \mathbf{k}_i}{\sqrt{d}} \right).
\end{equation}
Since each cluster is formed by assigning pruned tokens to their nearest anchors according to the feature similarity in \erahighlight{Sec.~\ref{ssec:recycling}}, it is naturally expected that tokens within the same cluster exhibit a local feature coherence.
Under this assumption (i.e., $\mathbf{k}_i \approx \mathbf{k}_k'$ for $i \in C_k$), we could approximate the sum by scaling the centroid contribution as follows:
\begin{equation}
    \resizebox{0.87\linewidth}{!}{$P_k^{\text{ideal}} \approx \sum_{i \in C_k} \exp\left( \frac{\mathbf{q}^\top \mathbf{k}_k^{\prime}}{\sqrt{d}} \right)
    = |C_k| \cdot \exp\left( \frac{\mathbf{q}^\top \mathbf{k}_k^{\prime}}{\sqrt{d}} \right)
    $}.
\end{equation}
However, the raw count $|C_k|$ assumes a uniform contribution from all tokens, which contradicts the saliency observations in \erahighlight{Sec.~\ref{ssec:recycling}}.
To account for varying token saliency, we replace $|C_k|$ with the attention logit bias $\mathbf{b}_k$.
Then, we can rectify the attention logit on the reduced sequence:
\begin{equation}
    P_k^{\text{rectified}}
    = \mathbf{b}_k \cdot \exp\left( \frac{\mathbf{q}^\top \mathbf{k}_k^{\prime}}{\sqrt{d}} \right).
\end{equation}
Equivalently, scaling the exponentiated logit by $\mathbf{b}_k$ can be realized by adding a log-bias term to the corresponding attention logit with the following equation:
\begin{equation}
    \log P_k^{\text{rectified}}
    =
    \frac{\mathbf{q}^\top \mathbf{k}_k^{\prime}}{\sqrt{d}}
    +
    \underbrace{\log \mathbf{b}_k}_{\text{Log-bias Term}}.
    \label{eq:lar_log_bias}
\end{equation}
With the above formulation, LAR ensures that the logit on the reduced token set $\mathbf{X}'$ reflects the contributions of all original tokens.
Meanwhile, to maintain the compatibility with optimized attention kernels~\cite{dao2022flashattention}, we propose a Hardware-aware Implementation (HI) that injects the log-bias term without modifying the kernel.
As shown in \erahighlight{Fig.~\ref{fig:overall} (d)}, we append one auxiliary dimension to the queries, keys, and values, carrying a constant query term, the scaled log-bias and a zero value term with the following equations:
\begin{equation}
    \tilde{\mathbf{Q}} = [\mathbf{Q}, \mathbf{1}], \quad
    \tilde{\mathbf{K}} = [\mathbf{K}, \sqrt{d}\mathbf{B}_{\log}], \quad
    \tilde{\mathbf{V}} = [\mathbf{V}, \mathbf{0}],
\end{equation}
where $\mathbf{B}_{\log}$ stores the element-wise $\log \mathbf{b}_k$.
Then, the dot-product logit naturally incorporates the bias as follows:
\begin{equation}
        \resizebox{0.80\linewidth}{!}{$
    \frac{\tilde{\mathbf{Q}}\tilde{\mathbf{K}}^\top}{\sqrt{d}}
    =
    \frac{\mathbf{Q}\mathbf{K}^\top + \mathbf{1}(\sqrt{d}\mathbf{B}_{\log})^\top}{\sqrt{d}}
    =
    \frac{\mathbf{Q}\mathbf{K}^\top}{\sqrt{d}} + \mathbf{B}_{\log}.
    $}
\end{equation}
Finally, after the attention operation, we discard the auxiliary padding, allowing LAR to implement the desired log-bias without changing the outputs.
In this way, LAR enables MLLMs to operate on the reduced sequence while preserving the original distribution for improved performance.
\\
\noindent\textbf{Theoretical Justification.}
The above implementation explains how LAR injects the bias without modifying attention kernels.
We now provide a theoretical derivation of the phenomenon we term \emph{Attention Logit Collapse} and establish a bounded rectification guarantee for LAR.
Specifically, for a cluster $C_k$ with $m=|C_k|$, let $\mathbf{k}_i$ denote the original key of token $i$.
We consider the local centroid model $\mathbf{k}_k'=\frac{1}{m}\sum_{i \in C_k}\mathbf{k}_i$, which matches the recycled representation when tokens in the cluster are locally coherent.
Given a query $\mathbf{q}$ and head dimension $d$, we can get the ideal attention logit in the original sequence and the naive attention logit after centroid update with the following equations:
\begin{equation}
    P_k^{\text{ideal}}
    =
    \sum_{i \in C_k}
    \exp\!\left(\frac{\mathbf{q}^{\top}\mathbf{k}_i}{\sqrt{d}}\right),
    \quad
    P_k^{\text{naive}}
    =
    \exp\!\left(\frac{\mathbf{q}^{\top}\mathbf{k}_k'}{\sqrt{d}}\right).
    \label{eq:ideal_naive_logit}
\end{equation}
Here, $P_k^{\text{ideal}}$ accumulates the contributions of original tokens in the cluster, whereas $P_k^{\text{naive}}$ keeps the centroid contribution.
Since $\exp(\cdot)$ is convex, Jensen's inequality~\cite{jensen1906fonctions} gives:
\begin{equation}
\begin{aligned}
    P_k^{\text{naive}}
    & =
    \exp\!\left(
    \frac{1}{m}\sum_{i \in C_k}
    \frac{\mathbf{q}^{\top}\mathbf{k}_i}{\sqrt{d}}
    \right)
    \leq
    \frac{1}{m}P_k^{\text{ideal}}, \\
    \frac{P_k^{\text{ideal}}}{P_k^{\text{naive}}}
    &\geq m
    \quad\Rightarrow\quad
    \text{\emph{Attention Logit Collapse}}.
\end{aligned}
    \label{eq:jensen_collapse}
\end{equation}
Thus, centroid replacement underestimates the cluster-level logit by at least a factor of $m$.
Therefore, visual evidence across multiple tokens collapses into a single logit after merging.
The cluster size is also a natural first-order target for rectification.
Let $\boldsymbol{\delta}_i=\mathbf{k}_i-\mathbf{k}_k'$ and assume $\|\boldsymbol{\delta}_i\|_2\leq \varepsilon$ for all $i \in C_k$.
Since $\mathbf{k}_k'$ is the centroid of the keys in $C_k$, the residuals satisfy $\sum_{i \in C_k}\boldsymbol{\delta}_i=\mathbf{0}$.
Thus, we have the following equation:
\begin{equation}
    \frac{P_k^{\text{ideal}}}{P_k^{\text{naive}}}
    =
    \sum_{i \in C_k}
    \exp\!\left(
    \frac{\mathbf{q}^{\top}\boldsymbol{\delta}_i}{\sqrt{d}}
    \right).
    \label{eq:ratio_residual}
\end{equation}
A Taylor expansion around zero cancels the first-order term after summing over the cluster.
With $\rho=\|\mathbf{q}\|_2\varepsilon/\sqrt{d}$, the remaining error is bounded with the second-order term:
\begin{equation}
    0
    \leq
    \frac{P_k^{\text{ideal}}}{P_k^{\text{naive}}}
    -
    m
    \leq
    \frac{1}{2}m e^{\rho}\rho^2
    =
    \mathcal{O}\!\left(
    \frac{m\|\mathbf{q}\|_2^2\varepsilon^2}{d}
    \right).
    \label{eq:second_order_bound}
\end{equation}
Therefore, the true multiplicative rectification is close to the cluster size, with only a second-order residual.
However, directly using the raw count $m$ treats all merged tokens as equally important.
ERA instead uses the saliency-aware bias $\mathbf{b}_k$ in Eq.~\eqref{eq:cluster_bias}.
From Eq.~\eqref{eq:bias_weight}, each pruned-token weight satisfies $0\leq \mathbf{w}_u\leq 1$, and thus $\mathbf{b}_k$ is bounded between one anchor contribution and the full cluster size.
Combining this bound with Eq.~\eqref{eq:jensen_collapse}, we obtain this sandwich formulation:
\begin{equation}
    1
    \leq
    \mathbf{b}_k
    \leq
    m
    \leq
    \frac{P_k^{\text{ideal}}}{P_k^{\text{naive}}},
    \qquad
    P_k^{\text{naive}}
    \leq
    P_k^{\text{rectified}}
    \leq
    P_k^{\text{ideal}} .
    \label{eq:lar_sandwich}
\end{equation}
The last inequality follows from $P_k^{\text{rectified}}=\mathbf{b}_kP_k^{\text{naive}}$.
This sandwich property shows that LAR moves the compressed attention logit toward the original one while avoiding over-amplification.
For highly salient clusters, $\mathbf{b}_k$ approaches $m$, providing a near-full compensation for the collapsed attention.
For heterogeneous clusters with less relevant tokens, $\mathbf{b}_k$ stays below $m$, yielding a conservative rectification that reflects token saliency rather than blindly restoring every merged token equally.
Overall, this derivation connects the three modules of ERA.
DEP selects anchors that support locally coherent clusters.
BTR converts entropy-derived saliency into a bounded cluster-level logit bias.
LAR injects this bias as an additive logit rectification within standard attention.
The below experiments verify that this rectification improves both accuracy and efficiency in practice.
\section{Experiments}
\subsection{Experimental Setup}
\label{ssec:setup}
We perform experiments to verify whether ERA preserves visual information, transfers across input regimes and yields practical inference gains under aggressive token reduction.
We first evaluate single-image understanding across diverse MLLM architectures, resolutions and benchmarks.
We then evaluate longer visual contexts, including multi-image and video understanding.
Finally, we analyze module contributions and deployment efficiency.
Unless otherwise specified, all experiments follow the official evaluation protocols.
\subsubsection{Models and Evaluation Scope}
\label{sssec:models}
We evaluate ERA on representative MLLMs that cover different visual-token generation paradigms.
\textbf{LLaVA-1.5}~\cite{liu2024improved} serves as the standard image-understanding backbone. It couples a CLIP~\cite{radford2021learning} vision encoder with a Vicuna~\cite{chiang2023vicuna} language model through an MLP projector, producing 576 visual tokens from $336 \times 336$ inputs.
\textbf{LLaVA-NeXT}~\cite{liu2024llavanext} represents the high-resolution setting. Dynamic grid pinning partitions each $672 \times 672$ image into multiple subimages aligned with the encoder's fixed resolution. These subimages are encoded independently and then concatenated, producing 2{,}880 visual tokens per image. This creates a substantially more severe redundancy problem.
For both families, we utilize the 7B and 13B variants to assess ERA's scalability across model capacity.
To verify architectural generality, we further evaluate \textbf{Qwen2.5-VL}~\cite{bai2025qwen2}. 
Its vision backbone builds on a Vision Transformer~\cite{dosovitskiy2021image} and incorporates RMSNorm~\cite{zhang2019root}, SwiGLU activations~\cite{shazeer2020glu} and window-based attention~\cite{liu2021swin}. 
An MLP-based Merger then fuses spatially adjacent patch features before feeding tokens into the LLM under dynamic-resolution inputs.
We then adopt \textbf{InternVL3}~\cite{zhu2025internvl3} as an additional strong MLLM to stress-test architectural robustness beyond the LLaVA and Qwen series.
Finally, we evaluate \textbf{NEO}~\cite{diao2025pixels}, a native MLLM that jointly processes visual and textual tokens in a single backbone. 
For longer visual contexts with multiple images, ERA is transferred to LLaVA-1.5 on MileBench~\cite{song2024milebench}. 
Beyond multi-image reasoning, we also apply ERA to \textbf{LLaVA-Video}~\cite{zhang2024video} and \textbf{LLaVA-OneVision}~\cite{li2024llavaonevision} for video understanding.
\begin{table*}[t]
    \centering
    \caption{
        Single-image comparison under \textbf{extreme compression} ($\sim$5\% tokens retained) with \textbf{LLaVA-1.5} and \textbf{LLaVA-NeXT} backbones. \textbf{Avg.} denotes retained performance relative to each full-token upper bound. Best and second-best pruned results are shown in \textbf{bold} and \underline{underlined}, respectively.
    }
    \resizebox{0.945\textwidth}{!}{
        \setlength{\tabcolsep}{6.5pt}
        \renewcommand{\arraystretch}{0.991}
        \begin{tabular}{lcccccccccc}
            \toprule
            \textbf{Method} & \textbf{VQA$^{\text{v2}}$} & \textbf{GQA} & \textbf{SQA$^{\text{I}}$} &
            \textbf{VQA$^{\text{T}}$} & \textbf{POPE} & \textbf{MME} &
            \textbf{MMB$^{\text{E}}$} & \textbf{MMB$^{\text{C}}$} &
            \textbf{MMVet} & \textbf{Avg. (\%)} \\
            \midrule

            % ================= LLaVA-1.5-7B =================
            \rowcolor{headergray}
            \multicolumn{11}{c}{\textsc{LLaVA-1.5-7B: Upper Bound: 576 Tokens (100\%)}} \\
            LLaVA-1.5-7B & 78.5 & 61.9 & 69.5 & 58.2 & 85.9 & 1506.5 & 64.7 & 58.1 & 31.3 & 100.0\% \\

            \rowcolor{headergray}
            \multicolumn{11}{c}{\textsc{Retain 32 Tokens} ($\downarrow$ 94.4\%)} \\
            PruMerge+\pub{(ICCV25)\cite{shang2025llava}}& 65.6 & 52.9 & 67.9 & 49.2 & 66.7 & 1236.6 & 55.1 & 45.9 & 24.7 & 83.8\% \\
            TRIM\pub{(COLING25)\cite{song2024trim}}    & 68.6 & 54.5 & 68.1 & 47.6 & 84.9 & 1251.8 & 57.7 & 40.1 & 20.5 & 84.5\% \\
            VisionZip\pub{(CVPR25)\cite{yang2025visionzip}} & 67.1 & 51.8 & 69.1 & 53.1 & 69.4 & 1251.2 & 57.0 & 50.3 & 25.3 & 86.6\% \\
            DART\pub{(EMNLP25)\cite{wen2025stop}}     & 67.1 & 52.9 & \underline{69.3} & 52.2 & 69.1 & 1273.3 & 58.5 & 50.0 & 25.0 & 86.9\% \\
            DivPrune\pub{(CVPR25)\cite{alvar2025divprune}}  & 71.2 & 54.9 & 68.6 & 52.9 & 81.5 & 1284.9 & 57.6 & 49.1 & 26.3 & 89.6\% \\
            CDPruner\pub{(NeurIPS25)\cite{zhang2025beyond}}  & \underline{73.6} & \textbf{57.0} & \textbf{69.5} & 53.2 & \textbf{87.9} & \textbf{1373.0} & 59.6 & 49.6 & \underline{27.8} & \underline{93.1\%} \\
            AgilePrune\pub{(ICLR26)\cite{baek2026agilepruner}} & \textbf{74.0} & 54.1 & 69.0 & \underline{54.5} & 80.1 & 1297.3 & \underline{60.4} & \textbf{53.6} & - & 92.4\% \\
            \rowcolor{rowblue}
            \textbf{Ours}       & \underline{73.6} & \underline{56.9} & 69.1 & \textbf{54.6} & \underline{85.6} & \underline{1312.6} & \textbf{60.9} & \underline{52.1} & \textbf{28.5} & \textbf{93.4\%} \\
            \midrule

            % ================= LLaVA-1.5-13B =================
            \rowcolor{headergray}
            \multicolumn{11}{c}{\textsc{LLaVA-1.5-13B: Upper Bound: 576 Tokens (100\%)}} \\
            LLaVA-1.5-13B & 80.0 & 63.3 & 72.8 & 61.2 & 86.0 & 1531.2 & 68.5 & 63.5 & 36.2 & 100.0\% \\
            
            \rowcolor{headergray}
            \multicolumn{11}{c}{\textsc{Retain 32 Tokens} ($\downarrow$ 94.4\%)} \\
            PruMerge+\pub{(ICCV25)\cite{shang2025llava}}& 66.8 & 54.1 & 71.7 & 52.4 & 67.4 & 1269.1 & 61.1 & 53.5 & 28.7 & 85.2\% \\
            TRIM\pub{(COLING25)\cite{song2024trim}}    & 69.8 & 55.6 & 70.4 & 49.6 & \underline{85.8} & 1284.7 & 63.1 & 45.4 & 26.4 & 85.9\% \\
            VisionZip\pub{(CVPR25)\cite{yang2025visionzip}} & 68.4 & 52.7 & \underline{72.9} & 55.2 & 66.8 & 1257.7 & 61.2 & 55.8 & 29.3 & 86.3\% \\
            DART\pub{(EMNLP25)\cite{wen2025stop}}     & 68.1 & 53.9 & \textbf{73.2} & 55.1 & 66.9 & 1282.8 & 61.9 & 56.2 & 29.4 & 86.9\% \\
            DivPrune\pub{(CVPR25)\cite{alvar2025divprune}}  & 72.0 & 56.2 & 70.9 & 54.6 & 79.3 & 1405.2 & 61.7 & \underline{57.2} & 27.8 & 89.6\% \\
            CDPruner\pub{(NeurIPS25)\cite{zhang2025beyond}} & \underline{75.2} & \textbf{58.5} & 71.9 & \underline{55.3} & \textbf{87.6} & \underline{1421.0} & \underline{63.7} & 56.6 & \underline{30.9} & \underline{93.0\%} \\
            \rowcolor{rowblue}
            \textbf{Ours}       & \textbf{75.4} & \underline{58.4} & 71.8 & \textbf{55.9} & 85.1 & \textbf{1461.4} & \textbf{65.8} & \textbf{59.4} & \textbf{32.6} & \textbf{94.5\%} \\
            \midrule

            % ================= LLaVA-NeXT-7B =================
            \rowcolor{headergray}
            \multicolumn{11}{c}{\textsc{LLaVA-NeXT-7B: Upper Bound: 2880 Tokens (100\%)}} \\
            LLaVA-NeXT-7B & 81.3 & 62.5 & 67.5 & 60.3 & 86.8 & 1511.8 & 65.8 & 57.3 & 40.0 & 100.0\% \\

            \rowcolor{headergray}
            \multicolumn{11}{c}{\textsc{Retain 160 Tokens} ($\downarrow$ 94.4\%)} \\
            PruMerge+\pub{(ICCV25)\cite{shang2025llava}}& 70.5 & 56.2 & 66.9 & 50.3 & 71.1 & 1289.6 & 58.0 & 48.9 & 29.3 & 85.9\% \\
            TRIM\pub{(COLING25)\cite{song2024trim}}    & 71.0 & 57.4 & 65.5 & 45.8 & 84.8 & 1275.8 & 61.6 & 45.2 & 29.6 & 86.7\% \\
            VisionZip\pub{(CVPR25)\cite{yang2025visionzip}} & 71.4 & 55.2 & \underline{67.9} & \underline{55.0} & 74.9 & 1327.8 & 58.6 & 50.4 & 32.3 & 88.9\% \\
            DART\pub{(EMNLP25)\cite{wen2025stop}}     & 72.5 & 56.8 & 67.8 & 54.9 & 75.3 & 1325.4 & 62.0 & 53.6 & 32.2 & 90.5\% \\
            DivPrune\pub{(CVPR25)\cite{alvar2025divprune}}  & 75.0 & 59.3 & 67.1 & 54.1 & 80.0 & 1356.6 & 62.9 & 53.7 & 32.0 & 91.9\% \\
            CDPruner\pub{(NeurIPS25)\cite{zhang2025beyond}} & \underline{76.7} & \textbf{60.8} & 67.5 & \textbf{55.4} & \textbf{86.8} & \textbf{1425.3} & \underline{64.2} & 53.8 & \textbf{36.2} & 95.5\% \\
            ZOO-Prune\pub{(CVPR26)\cite{kim2026zooprune}} & 76.1 & 59.9 & \textbf{68.4} & \textbf{55.4} & 83.1 & 1417.0 & \underline{64.2} & \underline{54.4} & - & \underline{95.6\%} \\
            \rowcolor{rowblue}
            \textbf{Ours}       & \textbf{77.3} & \underline{60.4} & \underline{67.9} & \underline{55.0} & \underline{85.2} & \underline{1422.2} & \textbf{64.5} & \textbf{57.1} & \underline{35.5} & \textbf{95.8\%} \\
            \midrule

            % ================= LLaVA-NeXT-13B =================
            \rowcolor{headergray}
            \multicolumn{11}{c}{\textsc{LLaVA-NeXT-13B: Upper Bound: 2880 Tokens (100\%)}} \\
            LLaVA-NeXT-13B & 82.3 & 64.4 & 73.1 & 63.2 & 85.3 & 1539.5 & 68.5 & 61.2 & 45.0 & 100.0\% \\

            \rowcolor{headergray}
            \multicolumn{11}{c}{\textsc{Retain 160 Tokens} ($\downarrow$ 94.4\%)} \\
            PruMerge+\pub{(ICCV25)\cite{shang2025llava}}& 71.6 & 57.9 & 70.1 & 52.8 & 72.1 & 1345.9 & 63.2 & 57.1 & 30.6 & 86.9\% \\
            TRIM\pub{(COLING25)\cite{song2024trim}}    & 72.1 & 58.9 & 69.1 & 49.2 & \underline{87.0} & 1392.3 & 65.7 & 51.6 & 27.8 & 87.3\% \\
            VisionZip\pub{(CVPR25)\cite{yang2025visionzip}} & 72.4 & 57.8 & 69.7 & \textbf{58.6} & 76.8 & 1393.9 & 64.8 & 60.0 & 35.9 & 91.0\% \\
            DART\pub{(EMNLP25)\cite{wen2025stop}}     & 72.8 & 58.7 & 70.1 & \underline{57.2} & 75.7 & 1389.3 & 64.6 & 60.8 & 35.0 & 90.7\% \\
            DivPrune\pub{(CVPR25)\cite{alvar2025divprune}}  & 75.6 & 60.0 & 71.4 & 56.3 & 81.9 & 1436.7 & 65.1 & \underline{60.9} & 37.4 & 93.2\% \\
            CDPruner\pub{(NeurIPS25)\cite{zhang2025beyond}} & \textbf{77.8} & \textbf{62.2} & \underline{71.7} & 56.7 & \textbf{88.3} & \underline{1476.9} & \underline{65.9} & 60.1 & \textbf{40.4} & \textbf{95.8\%} \\
            \rowcolor{rowblue}
            \textbf{Ours}       & \underline{77.3} & \textbf{62.2} & \textbf{72.3} & 56.3 & 82.6 & \textbf{1524.2} & \textbf{67.5} & \textbf{61.2} & \underline{37.8} & \underline{95.2\%} \\
            \bottomrule
        \end{tabular}
    }
    \label{tab:extreme_pruning_combined_all}
\end{table*}
\begin{table*}[t]
    \centering
    \caption{Token-budget comparison with \textbf{LLaVA-1.5-13B}.}
    \resizebox{0.945\textwidth}{!}{
        \setlength{\tabcolsep}{6.5pt}
        \renewcommand{\arraystretch}{1}

        \begin{tabular}{lcccccccccc}
            \toprule
            \textbf{Method} & \textbf{VQA$^{\text{v2}}$} & \textbf{GQA} & \textbf{SQA$^{\text{I}}$} &
            \textbf{VQA$^{\text{T}}$} & \textbf{POPE} & \textbf{MME} &
            \textbf{MMB$^{\text{E}}$} & \textbf{MMB$^{\text{C}}$} &
            \textbf{MMVet} & \textbf{Avg. (\%)} \\
            \midrule

            \rowcolor{headergray}
            \multicolumn{11}{c}{\textsc{Upper Bound: 576 Tokens (100\%)}} \\
            LLaVA-1.5-13B & 80.0 & 63.3 & 72.8 & 61.2 & 86.0 & 1531.2 & 68.5 & 63.5 & 36.2 & 100.0\% \\
            \midrule

            \rowcolor{headergray}
            \multicolumn{11}{c}{\textsc{Retain 128 Tokens} ($\downarrow$ 77.8\%)} \\
            FastV\pub{(ECCV24)\cite{chen2024image}}     & 75.3 & 58.3 & 74.2 & 58.6 & 75.5 & 1460.6 & 66.1 & 62.3 & 32.8 & 94.7\% \\
            PDrop\pub{(CVPR25)\cite{xing2024pyramid}}     & \textbf{78.2} & \textbf{61.0} & 73.3 & \textbf{60.2} & 83.6 & \textbf{1489.5} & 67.5 & \textbf{62.8} & 32.1 & 97.1\% \\
            SparseVLM\pub{(ICML25)\cite{zhang2024sparsevlm}} & 77.6 & 59.6 & \textbf{74.3} & \underline{59.3} & 85.0 & \underline{1487.9} & \textbf{68.4} & \underline{62.6} & 35.2 & 97.9\% \\
            PruMerge+\pub{(ICCV25)\cite{shang2025llava}}& 76.2 & 58.3 & 73.3 & 56.1 & 82.7 & 1445.9 & 66.3 & 61.2 & 33.6 & 95.1\% \\
            TRIM\pub{(COLING25)\cite{song2024trim}}    & 76.4 & 59.4 & 72.4 & 55.0 & 86.8 & 1426.9 & 67.1 & 58.4 & 35.1 & 95.5\% \\
            VisionZip\pub{(CVPR25)\cite{yang2025visionzip}} & 76.8 & 57.9 & 73.8 & 58.9 & 82.7 & 1449.2 & 67.4 & 62.5 & 36.0 & 96.9\% \\
            DART\pub{(EMNLP25)\cite{wen2025stop}}     & 75.7 & 57.7 & \underline{74.2} & 58.7 & 80.4 & 1395.0 & 65.4 & 62.2 & 34.8 & 95.3\% \\
            DivPrune\pub{(CVPR25)\cite{alvar2025divprune}}  & 77.1 & 59.2 & 72.8 & 58.0 & 86.8 & 1457.7 & 66.3 & 60.7 & 34.4 & 96.5\% \\
            VisPruner\pub{(ICCV25)\cite{wang2025vispruner}} & 76.9 & 58.4 & 73.9 & 59.2 & 83.8 & 1428.0 & 67.2 & 62.2 & 36.9 & 97.3\% \\
            CDPruner\pub{(NeurIPS25)\cite{zhang2025beyond}} & 77.7 & \underline{59.7} & 73.2 & 58.4 & \textbf{87.3} & 1478.0 & 67.5 & 61.5 & \underline{36.2} & \underline{97.9\%} \\
            AgilePrune\pub{(ICLR26)\cite{baek2026agilepruner}} & 77.5 & 59.1 & 72.8 & 58.9 & 86.9 & 1481.0 & \underline{67.6} & 61.9 & - & 97.6\% \\
            ZOO-Prune\pub{(CVPR26)\cite{kim2026zooprune}} & 77.8 & 58.9 & 73.4 & 58.8 & \underline{87.0} & 1484.7 & 67.0 & 60.9 & - & 97.4\% \\
            \rowcolor{rowblue}
            \textbf{Ours}       & \underline{77.9} & 59.5 & 73.4 & 58.2 & 86.9 & 1467.1 & 67.1 & 61.7 & \textbf{37.2} & \textbf{98.0\%} \\
            \midrule

            \rowcolor{headergray}
            \multicolumn{11}{c}{\textsc{Retain 64 Tokens} ($\downarrow$ 88.9\%)} \\
            FastV\pub{(ECCV24)\cite{chen2024image}}     & 65.3 & 51.9 & 73.1 & 53.4 & 56.9 & 1246.4 & 59.2 & 55.1 & 26.9 & 82.9\% \\
            PDrop\pub{(CVPR25)\cite{xing2024pyramid}}     & 70.8 & 54.1 & 73.1 & 55.3 & 66.1 & 1247.0 & 63.1 & 56.6 & 21.9 & 85.0\% \\
            SparseVLM\pub{(ICML25)\cite{zhang2024sparsevlm}} & 73.2 & 55.9 & 73.0 & 57.1 & 77.9 & 1374.3 & 65.2 & 60.3 & 32.9 & 92.8\% \\
            PruMerge+\pub{(ICCV25)\cite{shang2025llava}}& 72.6 & 56.3 & 73.5 & 54.4 & 75.7 & 1338.2 & 65.0 & 59.3 & 30.3 & 90.8\% \\
            TRIM\pub{(COLING25)\cite{song2024trim}}    & 73.2 & 57.9 & 72.0 & 52.0 & \underline{86.5} & 1406.2 & 65.0 & 52.7 & 27.8 & 90.4\% \\
            VisionZip\pub{(CVPR25)\cite{yang2025visionzip}} & 73.7 & 56.2 & \textbf{74.2} & 57.4 & 75.7 & 1379.6 & 64.9 & \underline{61.3} & 33.4 & 93.1\% \\
            DART\pub{(EMNLP25)\cite{wen2025stop}}     & 72.4 & 55.7 & \underline{73.8} & 57.3 & 72.8 & 1380.0 & 64.7 & 60.6 & 32.8 & 92.1\% \\
            DivPrune\pub{(CVPR25)\cite{alvar2025divprune}}  & 75.2 & 57.9 & 71.7 & 57.4 & 84.5 & \underline{1454.2} & 64.1 & 59.8 & 29.3 & 93.3\% \\
            VisPruner\pub{(ICCV25)\cite{wang2025vispruner}} & 73.9 & 56.0 & 74.0 & 57.9 & 79.2 & 1408.9 & 65.0 & 59.9 & 33.1 & 93.5\% \\
            CDPruner\pub{(NeurIPS25)\cite{zhang2025beyond}} & \underline{76.7} & \textbf{59.4} & 72.5 & 57.6 & \textbf{87.1} & \textbf{1466.8} & 65.5 & 58.8 & \underline{35.2} & \underline{96.2\%} \\
            AgilePrune\pub{(ICLR26)\cite{baek2026agilepruner}} & 75.7 & 57.5 & 72.0 & \textbf{58.6} & 82.0 & 1437.0 & \textbf{66.2} & \textbf{61.6} & - & 95.4\% \\
            ZOO-Prune\pub{(CVPR26)\cite{kim2026zooprune}} & 76.4 & 58.6 & 72.1 & \textbf{58.6} & 85.3 & 1438.6 & 64.8 & 60.1 & - & 95.7\% \\
            \rowcolor{rowblue}
            \textbf{Ours}       & \textbf{77.2} & \underline{58.8} & 72.5 & \underline{57.7} & 85.5 & 1446.2 & \underline{66.1} & 60.7 & \textbf{37.1} & \textbf{96.9\%} \\
            \bottomrule
        \end{tabular}
    }
    \label{tab:cdpruner_t3_llava15_13b}
\end{table*}
\subsubsection{Benchmarks}
\label{sssec:benchmarks}
For single-image evaluation, we group the benchmarks by the visual information required by each task.
General visual question answering and reasoning are evaluated with VQA$^{\text{v2}}$~\cite{goyal2017making}, GQA~\cite{hudson2019gqa} and SQA$^{\text{I}}$~\cite{lu2022learn}. These benchmarks test whether compressed tokens preserve object-level semantics, commonsense cues and scientific reasoning information. We report results on the VQA$^{\text{v2}}$ test-dev split, the GQA balanced test-dev split and the SQA image subset.
Fine-grained text, chart and diagram understanding are evaluated with VQA$^{\text{T}}$~\cite{singh2019towards}, ChartQA~\cite{masry2022chartqa} and AI2D~\cite{kembhavi2016diagram}. These benchmarks depend on local visual details, embedded text and layout structure. We evaluate them on the official validation, test and test splits, respectively.
The comprehensive multimodal capability is measured by MME~\cite{fu2025mme}, MMB$^{\text{E}}$~\cite{liu2024mmbench}, MMB$^{\text{C}}$~\cite{liu2024mmbench} and MMVet~\cite{yu2023mm}. Together, they cover perception, OCR, bilingual multiple-choice reasoning under CircularEval and open-ended LLM-judged multimodal assessment.
The object hallucination analysis is evaluated with POPE~\cite{li2023evaluating} on COCO~\cite{lin2014microsoft} and HaluBench~\cite{guan2024hallusionbench}. These benchmarks probe whether pruning methods retain the information for object presence and visual contradiction checking.
For multi-image reasoning, MileBench~\cite{song2024milebench} measures cross-image understanding, retrieval-oriented reasoning and long-context aggregation.
For video understanding, we follow the frame-based settings on VideoMME~\cite{fu2024videomme}, LongVideoBench~\cite{wu2024longvideobench} and MVBench~\cite{li2024mvbench}. 
These benchmarks extend the evaluation from spatial redundancy to repeated temporal context.
\subsubsection{Implementation Details}
\label{sssec:impl}
On image benchmarks, for the LLaVA-1.5 and LLaVA-NeXT series, we use the official evaluation implementations.
For Qwen2.5-VL, InternVL3 and NEO, we adopt VLMEvalKit~\cite{duan2024vlmevalkit} to ensure consistent and reproducible evaluation.
On video benchmarks, for LLaVA-Video and LLaVA-OneVision, we use lmms-eval~\cite{zhang2025lmmseval} to follow the standard video evaluation protocol.
All evaluations are implemented in PyTorch~\cite{paszke2019pytorch} and conducted on a single NVIDIA A800 GPU. The same setup is used for task performance and efficiency analysis, including latency measurement.
For Qwen2.5-VL, window-based attention is used in most vision blocks. Only the last vision block performs global attention over the full visual token sequence.
We therefore extract the attention map from this last global-attention block to ensure that the head-wise entropy is computed from globally comparable interactions rather than from window-restricted attention.
Since Qwen2.5-VL has no CLS token, we adopt a CLS-free formulation. We average attention over all query positions per head before entropy computation. 
To remain consistent with Qwen's MLP-Merger, we also average the resulting pre-merge entropy and auxiliary features within each merge group.
For a fair comparison, we adapt CDPruner~\cite{zhang2025beyond} to Qwen2.5-VL following its protocol for models without a text encoder. The text guidance is derived by averaging instruction-token embeddings, replacing the original CLIP-style pooling.
For InternVL3, we use the default CLS-token attention readout adopted by the LLaVA series. 
No architecture-specific adaptation is required because its vision encoder also provides a CLS token.
NEO does not expose a separate vision encoder. 
We therefore perform pruning during the prefill pass at the twelfth layer, the endpoint of the pre-Buffer stage. 
Meanwhile, we extract the submatrix of visual tokens from the full attention map and average it over all query positions per head before entropy computation.
Across all models, we set the bias-floor parameter in Eq.~\eqref{eq:bias_weight} to $\lambda=0.3$.
For other hyperparameters, we set the balance factor $\alpha=0.5$ and the regularization weight $\eta=0.1$ in all experiments unless otherwise specified.
\begin{table*}[t]
    \centering
    \caption{
    Multi-image comparison on \textbf{MileBench} with \textbf{LLaVA-1.5-7B}.
    }
    \label{tab:extra_milebench}
    \resizebox{0.945\textwidth}{!}{
        \setlength{\tabcolsep}{6.5pt}
        \renewcommand{\arraystretch}{1.05}

        \begin{tabular}{l|cccc|ccccc|cc|cc}
            \toprule
            \multirow{2}{*}{\textbf{Method}} &
            \multicolumn{4}{c|}{\textbf{Temporal Multi-image}} &
            \multicolumn{5}{c|}{\textbf{Semantic Multi-image}} &
            \multicolumn{2}{c|}{\textbf{Diagnostic Retrieval}} &
            \multicolumn{2}{c}{\textbf{Average}} \\
            \cmidrule(lr){2-5}
            \cmidrule(lr){6-10}
            \cmidrule(lr){11-12}
            \cmidrule(lr){13-14}
            &
            \textbf{T-1} & \textbf{T-2} & \textbf{T-3} & \textbf{T-4} &
            \textbf{S-1} & \textbf{S-2} & \textbf{S-3} & \textbf{S-4} & \textbf{S-5} &
            \textbf{NH} & \textbf{IR} &
            \textbf{Score} & \textbf{\%} \\
            \midrule

            \rowcolor{headergray}
            \multicolumn{14}{c}{\textsc{Upper Bound: Full Cache (100\%)}} \\

            LLaVA-1.5-7B &
            40.3 & 45.6 & 31.8 & 37.9 &
            56.9 & 33.2 & 12.7 & 24.3 & 61.0 &
            5.3 & 4.0 &
            32.1 & 100.0\% \\

            \midrule

            \rowcolor{headergray}
            \multicolumn{14}{c}{\textsc{Retain 10\% Tokens}} \\

            H2O\pub{(NeurIPS23)\cite{zhang2023h2o}} &
            37.4 & 42.9 & 29.0 & 35.1 &
            53.8 & 29.5 & 9.8 & 21.0 & 58.1 &
            2.8 & 1.5 &
            29.2 & 90.9\% \\

            SnapKV\pub{(NeurIPS24)\cite{li2024snapkv}} &
            36.9 & 42.4 & 28.6 & 34.6 &
            53.4 & 29.7 & 9.5 & 21.2 & 57.4 &
            1.9 & 1.7 &
            28.8 & 89.9\% \\

            PyramidKV\pub{(arXiv25)\cite{cai2024pyramidkv}} &
            37.1 & 42.7 & 28.8 & 34.9 &
            53.9 & 30.1 & 9.6 & 21.1 & 57.8 &
            2.1 & 1.4 &
            29.0 & 90.5\% \\

            LOOK-M\pub{(EMNLP24)\cite{wan2024lookm}} &
            38.8 & 44.0 & 30.0 & 36.3 &
            55.4 & 31.3 & 10.9 & 22.8 & 59.2 &
            3.5 & 2.6 &
            30.4 & 94.8\% \\

            MEDA\pub{(NAACL25)\cite{wan2025meda}} &
            40.1 & \textbf{45.1} & \textbf{31.1} & 37.4 &
            56.5 & 32.8 & 12.1 & 24.0 & 60.7 &
            4.9 & 4.1 &
            31.7 & 98.8\% \\

            FastV\pub{(ECCV24)\cite{chen2024image}} &
            32.0 & 42.4 & 28.2 & 34.2 &
            51.1 & 29.7 & \textbf{15.0} & 24.5 & 62.0 &
            5.3 & 3.2 &
            29.8 & 92.8\% \\

            VisionZip\pub{(CVPR25)\cite{yang2025visionzip}} &
            \textbf{44.7} & 25.9 & 20.8 & 39.1 &
            54.0 & 30.3 & 6.7 & \textbf{25.9} & 69.5 &
            12.2 & 9.7 &
            30.8 & 96.0\% \\

            \rowcolor{rowblue}
            \textbf{Ours} &
            42.3 & 33.8 & 18.0 & \textbf{43.9} &
            \textbf{58.5} & \textbf{34.8} & 7.9 & 24.6 & \textbf{71.0} &
            \textbf{12.5} & \textbf{11.3} &
            \textbf{32.6} & \textbf{101.6\%} \\

            \bottomrule
        \end{tabular}
    }
\end{table*}
\begin{table}[t]
    \centering
    \caption{
        Dynamic-resolution comparison with \textbf{Qwen2.5-VL-7B}.}
    \label{tab:qwen2.5_vl_7b_comparison_v3}
    
    % 请确保导言区定义了 \pub 命令
    % \newcommand{\pub}[1]{{\color{gray}\fontsize{6pt}{7pt}\selectfont #1}}
    
    \resizebox{0.489\textwidth}{!}{ 
        \setlength{\tabcolsep}{2.5pt} 
        \renewcommand{\arraystretch}{1.05} 

        \begin{tabular}{lccccccc}
            \toprule
            \textbf{Method} & \textbf{AI2D} & \textbf{ChartQA} & \textbf{POPE} & \textbf{MME} & \textbf{MMB$^{\text{E}}$} & \textbf{MMB$^{\text{C}}$} & \textbf{Avg. (\%)} \\
            \midrule

            % ================= UPPER BOUND =================
            \rowcolor{headergray}
            \multicolumn{8}{c}{\textsc{Upper Bound: Dynamic Resolution (100\%)}} \\
            Qwen2.5-VL-7B & 80.8 & 86.2 & 86.6 & 2306.5 & 79.8 & 82.4 & 100.0\% \\
            \midrule

            % ================= 50% TOKEN =================
            \rowcolor{headergray}
            \multicolumn{8}{c}{\textsc{Retain 50\% Tokens}} \\
            FastV\pub{(ECCV24)\cite{chen2024image}}       & \textbf{79.9} & \textbf{83.3} & \textbf{86.3} & 2286.7 & \textbf{80.2} & \underline{81.3} & \textbf{98.9\%} \\
            CDPruner\pub{(NeurIPS25)\cite{zhang2025beyond}} & 75.7 & 69.5 & \underline{85.7} & 2221.6 & 78.4 & 70.2 & 92.2\% \\
            VisionZip\pub{(CVPR25)\cite{yang2025visionzip}}   & 79.2 & 78.1 & 85.0 & \underline{2292.6} & \underline{80.1} & 81.0 & 97.5\% \\
            % HAWK\pub{(CVPR26)}\cite{zhu2026hawk} & 76.4 & 76.0 & 84.9 & 2245.3 & 78.7 & 68.9 & 90.8\% \\
            % DPPruner\pub{(NeurIPS25)}                  & 79.3 & 78.9 & 85.6 & 2257.1 & 79.1 & 72.5 & 95.6\% \\
            \rowcolor{rowblue}
            \textbf{Ours} & \underline{79.8} & \underline{82.7} & 84.9 & \textbf{2305.5} & 79.0 & \textbf{81.6} & \underline{98.5\%} \\
            \midrule

            % ================= 20% TOKEN =================
            \rowcolor{headergray}
            \multicolumn{8}{c}{\textsc{Retain 20\% Tokens}} \\
            FastV\pub{(ECCV24)\cite{chen2024image}}       & \underline{76.5} & \underline{57.8} & \underline{82.9} & \textbf{2225.4} & \textbf{78.8} & \underline{79.1} & \underline{91.5\%} \\
            CDPruner\pub{(NeurIPS25)\cite{zhang2025beyond}} & 70.8 & 52.0 & 80.4 & 2068.2 & 73.5 & 65.9 & 83.8\% \\
            VisionZip\pub{(CVPR25)\cite{yang2025visionzip}}   & 74.7 & 55.0 & 81.8 & 2212.2 & 78.1 & 78.6 & 90.0\% \\
            % DPPruner\pub{(NeurIPS25)}                  & 75.3 & 56.8 & 84.4 & 2200.1 & 75.8 & 65.9 & 87.8\% \\
            \rowcolor{rowblue}
            \textbf{Ours} & \textbf{77.2} & \textbf{70.2} & \textbf{83.4} & \underline{2224.3} & \underline{78.4} & \textbf{79.6} & \textbf{94.1\%} \\
            \midrule

            % ================= 10% TOKEN =================
            \rowcolor{headergray}
            \multicolumn{8}{c}{\textsc{Retain 10\% Tokens}} \\
            FastV\pub{(ECCV24)\cite{chen2024image}}       & \underline{72.6} & 34.6 & 75.8 & 1975.4 & 74.1 & 75.4 & 81.2\% \\
            CDPruner\pub{(NeurIPS25)\cite{zhang2025beyond}} & 67.9 & \underline{42.0} & 75.6 & 1872.3 & 69.0 & 62.9 & 77.3\% \\
            VisionZip\pub{(CVPR25)\cite{yang2025visionzip}}   & 69.6 & 38.3 & \underline{77.7} & \underline{2019.9} & \underline{75.2} & \underline{77.5} & \underline{82.7\%} \\
            % DPPruner\pub{(NeurIPS25)}                  & 71.8 & 40.0 & 81.9 & 1991.9 & 72.3 & 63.3 & 80.6\% \\
            \rowcolor{rowblue}
            \textbf{Ours} & \textbf{74.1} & \textbf{54.5} & \textbf{81.3} & \textbf{2041.2} & \textbf{77.1} & \textbf{78.5} & \textbf{88.2\%} \\
            \bottomrule
        \end{tabular}
    }
    \vspace{-3mm}
\end{table}
\begin{table}[t]
    \centering
    \caption{
        Image-domain comparison with \textbf{InternVL3-8B}.
    }
    \label{tab:extra_internvl3}
    
   \resizebox{0.489\textwidth}{!}{ 
        \setlength{\tabcolsep}{2.5pt} 
        \renewcommand{\arraystretch}{1.05} 
        
        \begin{tabular}{lcccccccc}
            \toprule
            \textbf{Method} & \textbf{VQA$^{\text{T}}$} & \textbf{MME} & \textbf{POPE} & \textbf{GQA} & \textbf{MMB$^{\text{E}}$} & \textbf{MMB$^{\text{C}}$} & \textbf{Avg. (\%)} \\
            \midrule
            
            % ================= UPPER BOUND =================
            \rowcolor{headergray}
            \multicolumn{8}{c}{\textsc{Upper Bound: 1,280 Tokens (100\%)}} \\
            InternVL3-8B & 81.5 & 2389.7 & 90.3 & 52.0 & 85.9 & 85.4 & 100.0\% \\
            \midrule
            
            % ================= 256 TOKENS =================
            \rowcolor{headergray}
            \multicolumn{8}{c}{\textsc{Retain 256 Tokens} ($\downarrow$ 80.0\%)} \\
            DivPrune\pub{(CVPR25)\cite{alvar2025divprune}} & \underline{67.1} & 2188.7 & \underline{90.1} & 49.0 & 82.0 & 80.8 & 93.0\% \\
            VisionZip\pub{(CVPR25)\cite{yang2025visionzip}} & 67.2 & 2192.0 & 88.6 & 47.8 & 82.9 & 81.5 & 92.7\% \\
            DART\pub{(EMNLP25)\cite{wen2025stop}} & 62.0 & \textbf{2280.4} & 88.0 & \textbf{50.9} & \textbf{83.2} & \textbf{83.2} & \underline{93.5\%} \\
            \rowcolor{rowblue}
            	\textbf{Ours} & \textbf{71.2} & \underline{2251.7} & \textbf{89.8} & \underline{48.7} & \underline{82.9} & \underline{83.0} & \textbf{94.7\%} \\
            \midrule
            
            % ================= 128 TOKENS =================
            \rowcolor{headergray}
            \multicolumn{8}{c}{\textsc{Retain 128 Tokens} ($\downarrow$ 90.0\%)} \\
            DivPrune\pub{(CVPR25)\cite{alvar2025divprune}} & 54.7 & 2043.0 & 87.6 & 47.4 & 78.6 & 78.0 & 87.3\% \\
            VisionZip\pub{(CVPR25)\cite{yang2025visionzip}} & 48.3 & 1987.8 & 85.2 & 44.6 & 77.7 & 76.5 & 83.8\% \\
            DART\pub{(EMNLP25)\cite{wen2025stop}} & 51.3 & \textbf{2180.4} & 85.1 & \textbf{48.9} & \textbf{80.7} & \textbf{80.9} & \underline{88.5\%} \\
            \rowcolor{rowblue}
            	\textbf{Ours} & \textbf{60.8} & \underline{2119.2} & \textbf{88.7} & \underline{47.7} & \underline{79.6} & \underline{79.8} & \textbf{89.9\%} \\
            \bottomrule
        \end{tabular}
    }
\end{table}

\begin{table}[t]
    \centering
    \caption{Encoder-free MLLM comparison with \textbf{NEO-2B}.}
    \label{tab:neo_2b_final_corrected}
    
    \resizebox{0.489\textwidth}{!}{ 
        \setlength{\tabcolsep}{2.5pt} 
        \renewcommand{\arraystretch}{1.1} 

        \begin{tabular}{lccccccc}
            \toprule
            \textbf{Method} & \textbf{AI2D} & \textbf{VQA$^{\text{T}}$} & \textbf{HaluBench} & \textbf{POPE} & \textbf{MMB$^{\text{E}}$} & \textbf{MMB$^{\text{C}}$} & \textbf{Avg. (\%)} \\
            \midrule

            % ================= UPPER BOUND =================
            \rowcolor{headergray}
            \multicolumn{8}{c}{\textsc{Upper Bound: Dynamic Resolution (100\%)}} \\
            NEO-2B      & 80.2 & 73.8 & 44.6 & 87.5 & 75.3 & 72.4 & 100.0\% \\
            \midrule

            % ================= 50% TOKEN =================
            \rowcolor{headergray}
            \multicolumn{8}{c}{\textsc{Retain 50\% Tokens}} \\
            FastV\pub{(ECCV24)\cite{chen2024image}}       & 78.1 & 53.7 & \textbf{43.4} & \underline{86.9} & \underline{75.4} & \textbf{71.8} & 94.4\% \\
            CDPruner\pub{(NeurIPS25)\cite{zhang2025beyond}} & 78.3 & 71.2 & 41.6 & 86.3 & 75.3 & \underline{71.5} & \underline{97.5\%} \\
            DivPrune\pub{(CVPR25)\cite{alvar2025divprune}}    & \underline{79.0} & \underline{67.2} & 42.0 & 86.3 & 73.6 & 70.8 & 96.3\% \\
            \rowcolor{rowblue}
            \textbf{Ours} & \textbf{79.2} & \textbf{72.4} & \underline{42.1} & \textbf{87.7} & \textbf{75.8} & 71.1 & \textbf{98.4\%} \\
            \midrule

            % ================= 20% TOKEN =================
            \rowcolor{headergray}
            \multicolumn{8}{c}{\textsc{Retain 20\% Tokens}} \\
            FastV\pub{(ECCV24)\cite{chen2024image}}       & 74.5 & 44.9 & \textbf{38.4} & 81.6 & \underline{72.0} & \underline{68.5} & 87.2\% \\
            CDPruner\pub{(NeurIPS25)\cite{zhang2025beyond}} & \underline{74.7} & \underline{62.2} & 36.9 & \underline{82.5} & 70.4 & 65.7 & \underline{89.8\%} \\
            DivPrune\pub{(CVPR25)\cite{alvar2025divprune}}    & 73.3 & 41.1 & 34.0 & 80.0 & 67.6 & 63.8 & 82.1\% \\
            \rowcolor{rowblue}
            \textbf{Ours} & \textbf{77.2} & \textbf{68.0} & \underline{37.6} & \textbf{88.1} & \textbf{75.1} & \textbf{70.9} & \textbf{95.2\%} \\
            \midrule

            % ================= 10% TOKEN =================
            \rowcolor{headergray}
            \multicolumn{8}{c}{\textsc{Retain 10\% Tokens}} \\
            FastV\pub{(ECCV24)\cite{chen2024image}}       & 68.9 & 30.8 & \textbf{36.6} & 73.9 & \underline{67.1} & \underline{65.1} & 78.9\% \\
            CDPruner\pub{(NeurIPS25)\cite{zhang2025beyond}} & \underline{70.0} & \underline{52.9} & 35.3 & \underline{78.0} & 66.7 & 60.6 & \underline{83.2\%} \\
            DivPrune\pub{(CVPR25)\cite{alvar2025divprune}}    & 66.9 & 23.3 & 32.4 & 67.0 & 53.2 & 50.7 & 67.5\% \\
            \rowcolor{rowblue}
            \textbf{Ours} & \textbf{73.7} & \textbf{60.5} & \underline{35.2} & \textbf{87.9} & \textbf{73.8} & \textbf{68.8} & \textbf{91.1\%} \\
            \bottomrule
        \end{tabular}
    }
\end{table}
\subsection{Single-image Evaluation}
\label{ssec:single_image}
Single-image evaluation is the primary test of whether visual token reduction preserves the information required for perception, OCR, chart reasoning, grounding and multimodal instruction following.
We begin with the LLaVA family under extreme compression in Tab.~\ref{tab:extreme_pruning_combined_all}, retaining 32 tokens for LLaVA-1.5 and 160 tokens for LLaVA-NeXT.
We then evaluate token-budget robustness on LLaVA-1.5-13B and extend the comparison to Qwen2.5-VL-7B, InternVL3-8B and NEO-2B.
Together, these settings evaluate ERA across different compression levels, model sizes and architectures.
\noindent\textbf{Main Results on the LLaVA Family.}
We analyze the LLaVA series across four regimes covering compression strength, model capacity, input resolution and reasoning difficulty.
\\
\noindent\textit{(1) Effectiveness in Standard Compression Regimes.}
We first evaluate LLaVA-1.5-7B with 32 retained tokens.
ERA maintains an average retention of 93.4\%, with particularly strong performance on fine-grained perception tasks.
Notably, ERA outperforms CDPruner by \textbf{2.5\%} on MMB$^{\text{C}}$, showing stronger preservation of semantic anchors for reasoning.
\\
\noindent\textit{(2) Scalability with Increased Model Capacity.}
When scaling to LLaVA-1.5-13B, ERA maintains 94.5\% of the upper-bound performance, surpassing CDPruner by \textbf{1.5\%}.
By selecting informative tokens, ERA allows the model to operate with a compact visual context without losing critical information.
\\
\noindent\textit{(3) Robustness in High-Resolution Scenarios.}
On the LLaVA-NeXT series, ERA achieves the highest average performance of 95.8\% on LLaVA-NeXT-7B. It also outperforms CDPruner by \textbf{0.6\%} on VQA$^{\text{v2}}$, indicating effective redundancy filtering.
\\
\noindent\textit{(4) Robustness in Complex Reasoning.}
Finally, on LLaVA-NeXT-13B, ERA remains competitive on average and is especially strong on benchmarks requiring complex reasoning and domain knowledge.
On MMB$^{\text{E}}$, ERA reaches 67.5\%, surpassing the prior best performance by 1.6\%.
These results show that ERA preserves the discriminative details required for high-level understanding under extreme compression.
\\
\noindent\textbf{Token-budget Robustness on LLaVA-1.5-13B.}
We further evaluate ERA on the larger 13B backbone under more moderate budgets in Tab.~\ref{tab:cdpruner_t3_llava15_13b}.
Across all three budgets, ERA ranks among the top methods in averaged performance, indicating robust behavior under both aggressive and mild visual token reduction.
Under the 32-token setting in Tab.~\ref{tab:extreme_pruning_combined_all}, ERA achieves the best results on six of nine benchmarks, with clear gains on VQA$^{\text{v2}}$ and VQA$^{\text{T}}$.
On the fine-grained benchmark MMB$^{\text{C}}$, ERA reaches 59.4\%, compared with 56.6\% for CDPruner. This gain of 2.8\% suggests better preservation of semantically critical visual anchors under tight information bottlenecks.
As the budget increases to 64 and 128 tokens in Tab.~\ref{tab:cdpruner_t3_llava15_13b}, the performance recovers smoothly on individual benchmarks. 
ERA still ranks first in averaged retention at both budgets, reaching 96.9\% at 64 tokens and 98.0\% at 128 tokens.
Overall, ERA remains strong on fine-grained tasks across the entire range from 32 to 128 tokens, indicating a balanced preservation of local details and global context.
\\
\noindent\textbf{Generalization to Advanced MLLMs.}
We verify the architectural generalization of ERA on three MLLM backbones whose visual-token generation paradigms differ substantially from the LLaVA family.
We present the results in order of increasing architectural deviations.
We start with the dynamic-resolution Qwen2.5-VL, move to another image-domain InternVL3, and finally to the encoder-free NEO.
\\
\noindent\textit{(1) Dynamic-resolution Qwen2.5-VL-7B.}
As shown in Tab.~\ref{tab:qwen2.5_vl_7b_comparison_v3}, ERA remains robust under progressive token reduction.
At 50\% tokens, ERA retains 98.5\% of the full-token performance and reaches 82.7\% on ChartQA, compared with 69.5\% for CDPruner.
Under the 20\% budget, ERA improves ChartQA from 52.0\% to 70.2\% over CDPruner and raises averaged retention from 83.8\% to 94.1\%.
These results show that ERA remains effective under dynamic-resolution inputs, particularly for OCR and reasoning sensitive to layout.
\\
\noindent\textit{(2) Image-domain InternVL3-8B.}
InternVL3-8B provides an additional strong image-domain backbone for testing architectural robustness.
As shown in Tab.~\ref{tab:extra_internvl3}, ERA achieves better performance than VisionZip under both 256 and 128 token settings.
Compared with DivPrune, ERA improves the performance across all benchmarks at both budgets, with clear gains on VQA$^{\text{T}}$ and MME.
These results demonstrate that ERA can effectively generalize to InternVL3.
\\
\noindent\textit{(3) Encoder-free NEO-2B.}
ERA generalizes consistently to NEO-2B as shown in Tab.~\ref{tab:neo_2b_final_corrected}, achieving 98.4\% retention at 50\% tokens.
On fine-grained tasks, ERA attains 60.5\% on VQA$^{\text{T}}$ and 73.7\% on AI2D at 10\% tokens.
ERA also outperforms CDPruner on VQA$^{\text{T}}$, improving the score from 52.9\% to 60.5\%. On POPE, the score increases from 78.0\% to 87.9\%.
These results support the generalization of ERA on advanced architectures beyond LLaVA-style MLLMs.
\begin{table*}[t]
    \centering
    \caption{
    Video comparison with \textbf{LLaVA-OneVision-7B} and \textbf{LLaVA-Video-7B}. \textsuperscript{I} denotes image-only pruning transferred to video inputs. Superscripts $\dagger$ and $\ddagger$ mark the best results within image-transfer and video-specific methods, respectively.
    }
    \label{tab:extra_llava_video}
    \resizebox{0.945\textwidth}{!}{
        \setlength{\tabcolsep}{6.5pt}
        \renewcommand{\arraystretch}{1.05}
        \begin{tabular}{l|cccc|cc|cc}
            \toprule
            \multirow{2}{*}{\textbf{Method}} &
            \multicolumn{4}{c|}{\textbf{VideoMME}} &
            \multirow{2}{*}{\makecell{\textbf{LongVideoBench}}} &
            \multirow{2}{*}{\textbf{MVBench}} &
            \multicolumn{2}{c}{\textbf{Avg. Acc.}} \\
            \cmidrule(lr){2-5}
            \cmidrule(lr){8-9}
            &
            \textbf{Short} &
            \textbf{Medium} &
            \textbf{Long} &
            \textbf{Overall} &
            &
            &
            \textbf{Score} &
            \textbf{\%} \\
            \midrule
            \rowcolor{headergray}
            \multicolumn{9}{c}{\textsc{LLaVA-OneVision-7B: Upper Bound (100\% Tokens)}} \\
            LLaVA-OneVision-7B & 69.9 & 56.7 & 48.9 & 58.5 & 56.6 & 58.3 & 57.8 & 100.0\% \\
            \midrule
            \rowcolor{headergray}
            \multicolumn{9}{c}{\textsc{Retain 25\% Tokens}} \\
            \rowcolor{rowblue}
            FastV\pub{(ECCV24)\cite{chen2024image}} & 68.1 & 54.7 & 46.8 & 56.5 & 55.4 & 56.4 & 56.1 & 97.1\% \\
            \rowcolor{rowblue}
            VisionZip\pub{(CVPR25)\cite{yang2025visionzip}} & 68.8 & 57.3$^\dagger$ & 48.2 & 58.1 & 56.4$^\dagger$ & 57.8$^\dagger$ & 57.4$^\dagger$ & 99.3\%$^\dagger$ \\
            \rowcolor{rowblue}
            Ours\textsuperscript{I} & 71.0$^\dagger$ & 56.6 & 49.1$^\dagger$ & 58.9$^\dagger$ & 55.6 & 57.0 & 57.2 & 99.0\% \\
            \rowcolor{rowgreen}
            PruneVID\pub{(ACL25)\cite{huang2024prunevid}} & 67.3 & 54.8 & 47.2 & 56.4 & 55.4 & 56.8 & 56.2 & 97.2\% \\
            \rowcolor{rowgreen}
            FastVID\pub{(NeurIPS25)\cite{shen2025fastvid}} & 69.9 & 56.3 & 47.4 & 57.9 & 55.9 & 58.1$^\ddagger$ & 57.3 & 99.1\% \\
            \rowcolor{rowgreen}
            FlashVID\pub{(ICLR26)\cite{fan2026flashvid}} & 71.2$^\ddagger$ & 57.0$^\ddagger$ & 49.3$^\ddagger$ & 59.2$^\ddagger$ & 56.8$^\ddagger$ & 58.0 & 58.0$^\ddagger$ & 100.3\%$^\ddagger$ \\
            \midrule
            \rowcolor{headergray}
            \multicolumn{9}{c}{\textsc{Retain 15\% Tokens}} \\
            \rowcolor{rowblue}
            FastV\pub{(ECCV24)\cite{chen2024image}} & 64.6 & 54.0 & 45.3 & 54.6 & 54.8 & 55.0 & 54.8 & 94.8\% \\
            \rowcolor{rowblue}
            VisionZip\pub{(CVPR25)\cite{yang2025visionzip}} & 63.8 & 54.6 & 48.3$^\dagger$ & 55.6 & 54.1 & 53.5 & 54.4 & 94.1\% \\
            \rowcolor{rowblue}
            Ours\textsuperscript{I} & 68.3$^\dagger$ & 56.4$^\dagger$ & 46.9 & 57.2$^\dagger$ & 55.0$^\dagger$ & 56.5$^\dagger$ & 56.2$^\dagger$ & 97.3\%$^\dagger$ \\
            \rowcolor{rowgreen}
            PruneVID\pub{(ACL25)\cite{huang2024prunevid}} & 67.2 & 52.8 & 46.7 & 56.1 & 54.5 & 55.1 & 55.2 & 95.6\% \\
            \rowcolor{rowgreen}
            FastVID\pub{(NeurIPS25)\cite{shen2025fastvid}} & 69.7$^\ddagger$ & 55.8 & 47.7 & 57.7 & 56.7 & 58.2$^\ddagger$ & 57.5 & 99.5\% \\
            \rowcolor{rowgreen}
            FlashVID\pub{(ICLR26)\cite{fan2026flashvid}} & 69.6 & 56.0$^\ddagger$ & 48.9$^\ddagger$ & 58.2$^\ddagger$ & 57.5$^\ddagger$ & 57.9 & 57.9$^\ddagger$ & 100.1\%$^\ddagger$ \\
            \midrule
            \rowcolor{headergray}
            \multicolumn{9}{c}{\textsc{Retain 10\% Tokens}} \\
            \rowcolor{rowblue}
            FastV\pub{(ECCV24)\cite{chen2024image}} & 60.9 & 52.2 & 44.9 & 52.7 & 52.4 & 53.4 & 52.8 & 91.3\% \\
            \rowcolor{rowblue}
            VisionZip\pub{(CVPR25)\cite{yang2025visionzip}} & 60.3 & 52.9 & 46.7 & 53.3 & 49.4 & 54.8 & 52.5 & 90.8\% \\
            \rowcolor{rowblue}
            Ours\textsuperscript{I} & 66.8$^\dagger$ & 53.7$^\dagger$ & 48.1$^\dagger$ & 56.2$^\dagger$ & 52.5$^\dagger$ & 56.0$^\dagger$ & 54.9$^\dagger$ & 95.0\%$^\dagger$ \\
            \rowcolor{rowgreen}
            PruneVID\pub{(ACL25)\cite{huang2024prunevid}} & 65.9 & 52.8 & 45.6 & 54.7 & 54.0 & 53.7 & 54.1 & 93.7\% \\
            \rowcolor{rowgreen}
            FastVID\pub{(NeurIPS25)\cite{shen2025fastvid}} & 68.1$^\ddagger$ & 55.7 & 47.8 & 57.2 & 55.7 & 57.0 & 56.6 & 98.0\% \\
            \rowcolor{rowgreen}
            FlashVID\pub{(ICLR26)\cite{fan2026flashvid}} & 67.3 & 57.1$^\ddagger$ & 49.0$^\ddagger$ & 57.8$^\ddagger$ & 56.5$^\ddagger$ & 57.4$^\ddagger$ & 57.2$^\ddagger$ & 99.0\%$^\ddagger$ \\
            \midrule
            \rowcolor{headergray}
            \multicolumn{9}{c}{\textsc{LLaVA-Video-7B: Upper Bound (100\% Tokens)}} \\
            LLaVA-Video-7B & 77.0 & 62.1 & 53.3 & 64.2 & 59.5 & 61.9 & 61.9 & 100.0\% \\
            \midrule
            \rowcolor{headergray}
            \multicolumn{9}{c}{\textsc{Retain 25\% Tokens}} \\
            \rowcolor{rowblue}
            FastV\pub{(ECCV24)\cite{chen2024image}} & 71.7 & 59.2 & 50.9 & 60.6 & 56.4 & 59.1 & 58.7 & 94.8\% \\
            \rowcolor{rowblue}
            VisionZip\pub{(CVPR25)\cite{yang2025visionzip}} & 74.0$^\dagger$ & 60.3$^\dagger$ & 52.9$^\dagger$ & 62.4$^\dagger$ & 58.3 & 60.0 & 60.2 & 97.3\% \\
            \rowcolor{rowblue}
            Ours\textsuperscript{I} & 72.3 & 57.0 & 51.6 & 60.3 & 60.2$^\dagger$ & 60.5$^\dagger$ & 60.3$^\dagger$ & 97.5\%$^\dagger$ \\
            \rowcolor{rowgreen}
            FastVID\pub{(NeurIPS25)\cite{shen2025fastvid}} & 74.7$^\ddagger$ & 60.1 & 53.6$^\ddagger$ & 62.8$^\ddagger$ & 58.2 & 60.5$^\ddagger$ & 60.5 & 97.7\% \\
            \rowcolor{rowgreen}
            FlashVID\pub{(ICLR26)\cite{fan2026flashvid}} & 74.2 & 61.4$^\ddagger$ & 51.6 & 62.4 & 59.1$^\ddagger$ & 60.2 & 60.6$^\ddagger$ & 97.8\%$^\ddagger$ \\
            \midrule
            \rowcolor{headergray}
            \multicolumn{9}{c}{\textsc{Retain 10\% Tokens}} \\
            \rowcolor{rowblue}
            FastV\pub{(ECCV24)\cite{chen2024image}} & 64.3 & 53.8 & 49.2 & 55.8 & 53.6 & 56.2 & 55.2 & 89.2\% \\
            \rowcolor{rowblue}
            VisionZip\pub{(CVPR25)\cite{yang2025visionzip}} & 69.4 & 57.9$^\dagger$ & 51.2$^\dagger$ & 59.5$^\dagger$ & 54.5 & 58.5$^\dagger$ & 57.5 & 92.9\% \\
            \rowcolor{rowblue}
            Ours\textsuperscript{I} & 69.9$^\dagger$ & 56.8 & 49.4 & 58.7 & 56.4$^\dagger$ & 57.7 & 57.6$^\dagger$ & 93.1\%$^\dagger$ \\
            \rowcolor{rowgreen}
            FastVID\pub{(NeurIPS25)\cite{shen2025fastvid}} & 71.8 & 57.3 & 50.2 & 59.8 & 56.9 & 59.3$^\ddagger$ & 58.7 & 94.8\% \\
            \rowcolor{rowgreen}
            FlashVID\pub{(ICLR26)\cite{fan2026flashvid}} & 72.2$^\ddagger$ & 59.1$^\ddagger$ & 51.2$^\ddagger$ & 60.9$^\ddagger$ & 57.7$^\ddagger$ & 59.3$^\ddagger$ & 59.3$^\ddagger$ & 95.8\%$^\ddagger$ \\
            \bottomrule
        \end{tabular}
    }
\end{table*}

\subsection{Multi-image Evaluation}
\label{ssec:multi_image}
We next evaluate multi-image reasoning, where redundancy is distributed across images.
This setting tests whether the retained tokens still support cross-image aggregation and retrieval-oriented reasoning after token compression.
\\
\noindent\textbf{Multi-image Long-context Results.}
We evaluate ERA on MileBench, where each method retains approximately 58 visual tokens per image. This budget corresponds to a 10\% retention ratio for LLaVA-1.5-7B.
As shown in Tab.~\ref{tab:extra_milebench}, ERA is competitive with both long-context cache-compression baselines and transferable image-token reduction baselines in multi-image scenarios.
ERA achieves the best average score, reaching 32.6\% and retaining 101.6\% of the full-cache upper bound.
Compared with MEDA, ERA improves the average score from 31.7\% to 32.6\% and performs better in 8 out of 11 categories.
The gains are most pronounced on T-4, S-5, NH and IR.
These experiments demonstrate that ERA can effectively handle multi-image long-context reasoning.
\subsection{Video Evaluation}
\label{ssec:video}
We further evaluate ERA on video understanding tasks.
%
% Although our proposed ERA is specifically designed for image-token reduction, video MLLMs also process long visual sequences when many frames are encoded jointly.
%
This setting examines whether ERA can extend to video inputs without introducing temporal-specific modules.
\\
\noindent\textbf{Video Understanding Results.}
Across video MLLMs, ERA remains competitive with general image-pruning baselines such as FastV and VisionZip.
With LLaVA-Video-7B at 25\% retention, ERA reaches 60.2\% on LongVideoBench and 60.5\% on MVBench. The latter matches FastVID under the same token budget. Its average accuracy reaches 60.3\%, corresponding to 97.5\% of the full-token model.
Under the same 25\% video setting, a separate LAR ablation with LLaVA-Video-7B shows that adding LAR yields 60.2\% on LongVideoBench, improving the accuracy by 1.8\% and confirming the effectiveness of attention rectification under long-context visual inputs.
At around 10\% retention, ERA maintains competitive performance on VideoMME. 
With LLaVA-Video-7B, it obtains 57.7\% on MVBench and an average accuracy of 57.6\%.
Within transferable image-pruning methods, ERA achieves the best average accuracy and the strongest result on LongVideoBench.
These results show that our proposed modules transfer effectively to video long-context inference, remaining competitive with video-specific methods without explicit temporal modeling.
\subsection{Module Effectiveness Verification}
\label{ssec:module}
In this section, we evaluate the effectiveness of ERA with detailed module ablations, readout stability analysis, layer-wise LAR analysis and attention-restoration verification.
\begin{table}[t]
    \centering
    \caption{
        Module ablation with \textbf{LLaVA-1.5-7B}. DEP($\mathbf{x}$) and DEP($\mathbf{y}$) denote diversity-based and saliency-based DEP, respectively.
    }
    \label{tab:ablation_main}
    
    \resizebox{0.489\textwidth}{!}{ 
        \setlength{\tabcolsep}{4.0pt}
        \small
        \renewcommand{\arraystretch}{1.05}

        \begin{tabular}{@{}cccccccc@{}}
            \toprule
            \multicolumn{4}{c}{\textbf{Modules}} & \multicolumn{4}{c}{\textbf{Downstream Tasks}} \\
            \cmidrule(lr){1-4} \cmidrule(lr){5-8}
            	\textbf{DEP($\mathbf{x}$)} & \textbf{DEP($\mathbf{y}$)} & \textbf{BTR} & \textbf{LAR} & \textbf{VQA$^{\text{T}}$} & \textbf{POPE} & \textbf{GQA} & \textbf{Avg. (\%)} \\
            \midrule

            % ================= UPPER BOUND =================
            \rowcolor{headergray}
            \multicolumn{8}{c}{\textsc{Upper Bound: 576 Tokens (100\%)}} \\
            \multicolumn{4}{c}{LLaVA-1.5-7B} & 58.2 & 85.9 & 61.9 & 100.0\% \\
            \midrule

            % ================= 128 TOKENS =================
            \rowcolor{headergray}
                 \multicolumn{8}{c}{\textsc{Retain 128 Tokens} ($\downarrow$ 77.8\%)} \\
                 \cmark &        &        &        & 55.8 & 85.4 & 59.0 & 96.9\% \\
                     & \cmark &        &        & \textbf{56.9} & 82.6 & 57.9 & 95.8\% \\
                 \cmark & \cmark &        &        & 56.3 & 87.0 & 59.2 & 97.9\% \\
                 \cmark & \cmark & \cmark &        & 56.5 & \underline{87.1} & \underline{59.2} & \underline{98.0\%} \\
            % Full Method Row
            \rowcolor{rowblue}
                  \cmark & \cmark & \cmark & \cmark & \underline{56.6} & \textbf{87.6} & \textbf{59.6} & \textbf{98.5\%} \\
            \midrule

            % ================= 64 TOKENS =================
            \rowcolor{headergray}
                 \multicolumn{8}{c}{\textsc{Retain 64 Tokens} ($\downarrow$ 88.9\%)} \\
                  \cmark &        &        &        & 54.1 & 84.1 & 57.5 & 94.6\% \\
                     & \cmark &        &        & 54.5 & 75.7 & 55.4 & 90.4\% \\
                 \cmark & \cmark &        &        & 55.2 & \underline{86.0} & 57.8 & 96.1\% \\
                 \cmark & \cmark & \cmark &        & \underline{55.7} & 85.2 & \underline{58.0} & \underline{96.2\%} \\
            % Full Method Row
            \rowcolor{rowblue}
                  \cmark & \cmark & \cmark & \cmark & \textbf{55.9} & \textbf{87.0} & \textbf{58.5} & \textbf{97.3\%} \\
            \midrule

            % ================= 32 TOKENS =================
            \rowcolor{headergray}
                 \multicolumn{8}{c}{\textsc{Retain 32 Tokens} ($\downarrow$ 94.4\%)} \\
                 \cmark &        &        &        & 52.9 & 81.8 & 54.2 & 91.2\% \\
                     & \cmark &        &        & 53.3 & 68.2 & 51.7 & 84.8\% \\
                 \cmark & \cmark &        &        & 53.7 & 82.9 & 55.3 & 92.7\% \\
                 \cmark & \cmark & \cmark &        & \underline{54.3} & \underline{83.1} & \underline{56.3} & \underline{93.7\%} \\
            % Full Method Row
            \rowcolor{rowblue}
                  \cmark & \cmark & \cmark & \cmark & \textbf{54.6} & \textbf{85.6} & \textbf{56.9} & \textbf{95.1\%} \\
            \bottomrule
        \end{tabular}
    }
\end{table}
\\
\noindent\textbf{Analysis of Key Modules.}
Tab.~\ref{tab:ablation_main} isolates the contribution of each component.
DEP($\mathbf{x}$) preserves representative visual patterns across compression ratios and achieves 94.6\% retention at 64 tokens.
DEP($\mathbf{y}$) drops to 90.4\%, indicating that saliency without diversity is insufficient to maintain global semantics.
Combining DEP($\mathbf{x}$) with DEP($\mathbf{y}$) consistently improves robustness, confirming the complementarity of visual diversity and entropy-guided saliency.
Introducing BTR further enhances performance under aggressive compression, improving retention by up to \textbf{1.0\%} at 32 tokens.
LAR increases the average retention from 93.7\% to \textbf{95.1\%}, effectively rectifying attention logit shifts caused by severe token reduction.
These results confirm the contribution of each module under varying compression budgets.
\begin{table}[!t]
    \centering
    \caption{
    CLS-free average-readout ablation on representative large-scale benchmarks. $\Delta$ rows report normalized changes.
    }
    \label{tab:extra_cls_free_common}
    \resizebox{\linewidth}{!}{
        \setlength{\tabcolsep}{3.8pt}
        \footnotesize
        \renewcommand{\arraystretch}{0.98}
        \begin{tabular}{@{}clccccc@{}}
            \toprule
            \textbf{Budget} & \textbf{Readout} & \textbf{GQA} & \textbf{SQA$^{\text{I}}$} & \textbf{VQA$^{\text{T}}$} & \textbf{POPE} & \textbf{Avg. (\%)} \\
            \midrule
            \rowcolor{headergray}
            \multicolumn{7}{c}{\textsc{LLaVA-1.5-7B}} \\
            Full model & Baseline & 61.9 & 69.5 & 58.2 & 85.9 & 100.0\% \\
            \multirow{3}{*}{Retain 64} & ERA-CLS & 58.5 & 68.8 & 55.9 & 87.0 & \textbf{97.7\%} \\
            & ERA-Avg & 58.3 & 68.2 & 55.5 & 87.4 & 97.4\% \\
            \rowcolor{rowblue}
            & $\Delta$ & -0.3\% & -0.9\% & -0.7\% & +0.5\% & -0.3\% \\
            \multirow{3}{*}{Retain 32} & ERA-CLS & 56.9 & 69.1 & 54.6 & 85.6 & \textbf{96.2\%} \\
            & ERA-Avg & 57.0 & 67.9 & 53.8 & 86.5 & 95.7\% \\
            \rowcolor{rowblue}
            & $\Delta$  & +0.2\% & -1.7\% & -1.4\% & +1.0\% & -0.5\% \\
            \midrule
            \rowcolor{headergray}
            \multicolumn{7}{c}{\textsc{LLaVA-NeXT-7B}} \\
            Full model & Baseline & 62.5 & 67.5 & 60.3 & 86.8 & 100.0\% \\
            \multirow{3}{*}{Retain 320} & ERA-CLS & 61.7 & 67.9 & 56.5 & 86.9 & 98.3\% \\
            & ERA-Avg & 61.8 & 68.4 & 56.3 & 86.7 & \textbf{98.4\%} \\
            \rowcolor{rowblue}
            & $\Delta$ & +0.2\% & +0.7\% & -0.3\% & -0.2\% & +0.1\% \\
            \multirow{3}{*}{Retain 160} & ERA-CLS & 60.4 & 67.9 & 55.0 & 85.2 & \textbf{96.7\%} \\
            & ERA-Avg & 60.7 & 67.8 & 54.8 & 83.9 & 96.3\% \\
            \rowcolor{rowblue}
            & $\Delta$ & +0.5\% & -0.1\% & -0.3\% & -1.5\% & -0.4\% \\
            \bottomrule
        \end{tabular}
    }
\end{table}
\\
\noindent\textbf{CLS and AVG Readout Stability.}
ERA uses head-wise attention entropy to estimate visual-token saliency. When the vision encoder provides a CLS token, the default implementation uses the CLS-token attention row as the saliency readout.
For architectures without a CLS token, this readout can be replaced by average pooling over all patch-token attention rows.
To verify that ERA does not depend on this design choice, we compare CLS and average readouts with the LLaVA-1.5-7B and LLaVA-NeXT-7B backbones in Tab.~\ref{tab:extra_cls_free_common}.
The close agreement with the LLaVA family indicates that ERA's entropy-based saliency is not inherently tied to the CLS token.
This conclusion is further supported by the results of CLS-free architectures.
For Qwen2.5-VL, ERA computes entropy from the last global-attention block and remains effective under dynamic-resolution inputs, as shown in Tab.~\ref{tab:qwen2.5_vl_7b_comparison_v3}.
For NEO-2B, which is encoder-free and has no vision-encoder CLS token, ERA still preserves strong performance in Tab.~\ref{tab:neo_2b_final_corrected}. These results show that the entropy readout can be adapted to the available attention structure rather than depending on a specific CLS-token design.
\begin{table}[!t]
    \centering
    \caption{
    LAR strategy ablation with \textbf{LLaVA-1.5-7B} and \textbf{LLaVA-NeXT-7B}.
    }
    \label{tab:extra_lar_layer_strategy}
    \resizebox{0.489\textwidth}{!}{
        \setlength{\tabcolsep}{4.0pt}
        \small
        \renewcommand{\arraystretch}{1.08}
        \begin{tabular}{@{}lcccc@{}}
            \toprule
            \multirow{2}{*}{\textbf{LAR Strategy}} &
            \multicolumn{1}{c@{}}{\textbf{LLaVA-1.5-7B}} &
            \multicolumn{3}{c@{}}{\textbf{LLaVA-NeXT-7B}} \\
            \cmidrule(l){2-5}
            &\textbf{VQA$^{\text{T}}$}  & \textbf{VQA$^{\text{T}}$} & \textbf{POPE} & \textbf{SQA$^{\text{I}}$} \\
            \midrule
            \rowcolor{rowblue}
            Default: uniform all layers & 57.01 & \textbf{56.50} & \textbf{86.65} & \textbf{68.03} \\
            Early only: layers 0, 1, and 2 & 57.02 & 56.35 & 86.43 & 67.98 \\
            Decayed: linear across layers & \textbf{57.03} & 56.39 & 86.40 & 68.00 \\
            \bottomrule
        \end{tabular}
    }
\end{table}

\begin{figure*}[!t]
    \centering
    \includegraphics[width=1\textwidth]{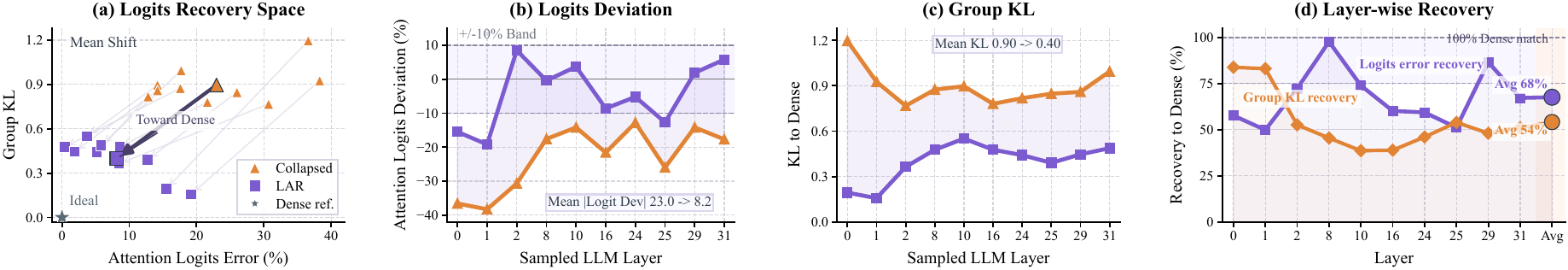}
    \caption{LAR verification with \textbf{LLaVA-1.5-7B} on VQA$^{\text{T}}$. (a) Joint trajectory of attention logit error and token-group KL divergence, where arrows indicate the shift from Collapsed to LAR toward the dense-reference distribution. (b) Signed attention logit deviation from the unpruned model. (c) Token-group KL divergence to the grouped unpruned attention distribution. (d) Layer-wise recovery of attention logit error and group-level KL.}
    \label{fig:extra_lar_mechanism}
\end{figure*}
\begin{table*}[t]
    \centering
    \caption{
        Efficiency and accuracy comparison with \textbf{LLaVA-NeXT-7B}. Metrics cover FLOPs, latency, memory usage and average performance retention.
        }
    \label{tab:efficiency_comparison_1_6_final}
    \resizebox{0.925\textwidth}{!}{
        \setlength{\tabcolsep}{4pt}
        \renewcommand{\arraystretch}{1.05}

        % 新增了一列，共9列
        \begin{tabular}{l | c c c c c c c c}
            \toprule
            \textbf{Method} & \textbf{Token} & \textbf{FLOPs} & \textbf{E2E Latency} & \textbf{Prefill} & \textbf{Decode / Token} & \textbf{KV Cache} & \textbf{GPU Memory} & \textbf{Avg.} \\
            & \scriptsize{(\#)} & \scriptsize{(TFLOPs)} & \scriptsize{(ms)} & \scriptsize{(ms)} & \scriptsize{(ms)} & \scriptsize{(GB)} & \scriptsize{(GB)} & \scriptsize{(\%)} \\
            \midrule

            % ================= BASELINE =================
            \rowcolor{headergray}
            \multicolumn{9}{c}{\textsc{Baseline: Full Model}} \\
            
            LLaVA-NeXT-7B & 2880 & 66.1 & 544 & 415 & 25 & 1.44 & 18.6 & 100.0\% \\
            \midrule

            % ================= COMPRESSED METHODS =================
            \rowcolor{headergray}
            \multicolumn{9}{c}{\textsc{Lightweight Methods (Retain 160 Tokens)}} \\
            PDrop\pub{(CVPR25)\cite{xing2024pyramid}}       & 160 & 17.4 & 249 & 129 & 29 & \textbf{0.08} & 16.1 & $<$79.2\% \\
            VisionZip\pub{(CVPR25)\cite{yang2025visionzip}}   & 160 & \textbf{5.6} & 232 & 93 & 33 & \textbf{0.08} & 15.3 & 88.9\% \\
            DART\pub{(EMNLP25)\cite{wen2025stop}}       & 160 & \underline{8.6} & \textbf{192} & 93 & \textbf{27} & \textbf{0.08} & \textbf{13.8} & 90.5\% \\
            CDPruner\pub{(NeurIPS25)\cite{zhang2025beyond}} & 160 & \textbf{5.6} & \underline{198} & \textbf{85} & \underline{28} & \textbf{0.08} & \underline{14.3} & \underline{95.5\%} \\
            
            \rowcolor{rowblue}
            \textbf{Ours} & 160 & \textbf{5.6} & 200 & \underline{91} & \textbf{27} & \textbf{0.08} & 14.6 & \textbf{95.8\%} \\
            \bottomrule
        \end{tabular}
    }
\end{table*}

\begin{table}[!t]
    \centering
    \caption{Efficiency and accuracy impact of LAR with \textbf{LLaVA-NeXT-7B}.}\label{tab:extra_lar_overhead}
    \resizebox{\linewidth}{!}{ 
        \setlength{\tabcolsep}{4.0pt}
        \small
        \renewcommand{\arraystretch}{1.05}
        \begin{tabular}{@{}lccc@{}}
            \toprule
            \textbf{Metric} & \textbf{w/o LAR} & \textbf{ERA} & \textbf{Gain} \\
            \midrule
            \rowcolor{headergray}
            \multicolumn{4}{c}{\textsc{Latency}} \\
            E2E Latency (ms) & 135.300 & 134.608 & \textbf{0.692 ms lower} \\
            LLM Prefill (ms) & 92.784 & 92.139 & 0.645 ms lower \\
            Decode Total (ms) & 42.516 & 42.469 & 0.047 ms lower \\
            Decode Throughput (tok/s) & 32.9 & 33.1 & 0.2 tok/s higher \\
            \midrule
            \rowcolor{headergray}
            \multicolumn{4}{c}{\textsc{Resource}} \\
            Peak GPU Memory (GB) & 14.633 & 14.633 & \textbf{unchanged} \\
            \midrule
            \rowcolor{headergray}
            \multicolumn{4}{c}{\textsc{Accuracy}} \\
            GQA Accuracy (\%) & 59.68 & \textbf{60.55} & \textbf{0.87\% higher} \\
            \bottomrule
        \end{tabular}
    }
\end{table}

\begin{table}[t]
    \centering
    \caption{
    vLLM serving benchmark with \textbf{LLaVA-NeXT-7B}.
    }
    \label{tab:extra_vllm_deployment_rowwise}
    \resizebox{\linewidth}{!}{
        {\small
        \setlength{\tabcolsep}{1.5pt}
        \renewcommand{\arraystretch}{1.05}
        \begin{tabular}{@{}lccc@{}}
            \toprule
            \textbf{Metric} & \textbf{Baseline} & \textbf{ERA} & \textbf{Gain} \\
            \midrule
            \rowcolor{headergray}
            \multicolumn{4}{c}{\textsc{Batch=256}} \\
            Amortized E2E Time / Req. (ms) & 77.7 & 23.9 & \textbf{3.3$\times$ speedup} \\
            LLM Prefill(ms) & 167.01 & 39.22 & \textbf{4.3$\times$ speedup} \\
            Decode Throughput (tok/s) & 68.3 & 70.4 & comparable \\
            Avg. Prefill Tokens & 1845 & 373 & \textbf{4.9$\times$ compression} \\
            KV Cache / Req. (GB) & 0.901 & 0.182 & \textbf{5.0$\times$ reduction} \\
            Peak GPU Memory (GB) & 243.69 & 59.69 & \textbf{4.1$\times$ lower} \\
            \midrule
            \rowcolor{headergray}
            \multicolumn{4}{c}{\textsc{Batch=48}} \\
            Amortized E2E Time / Req. (ms) & 81.8 & 37.6 & \textbf{2.2$\times$ speedup} \\
            LLM Prefill(ms) & 57.91 & 17.89 & \textbf{3.2$\times$ speedup} \\
            KV Cache / Req. (GB) & 0.272 & 0.046 & \textbf{5.9$\times$ reduction} \\
            Peak GPU Memory (GB) & 26.16 & 15.35 & \textbf{10.81 GB lower} \\
            \midrule
            \rowcolor{headergray}
            \multicolumn{4}{c}{\textsc{Batch=1}} \\
            Amortized E2E Time / Req. (ms) & 611.4 & 507.3 & \textbf{1.2$\times$ speedup} \\
            LLM Prefill(ms) & 37.02 & 15.73 & \textbf{2.4$\times$ speedup} \\
            \bottomrule
        \end{tabular}
        }
    }
\end{table}
\\
\noindent\textbf{LAR Strategy across Layers.}
We further examine whether LAR requires an explicit bias across layers.
Without explicit layer-wise scheduling, ERA stays within 0.25\% of the tested strategies, keeping LAR simple and efficient.
This confirms that LAR is not sensitive to layer-wise scheduling.
\\
\noindent\textbf{LAR Mechanism Verification.}
LAR is designed to correct the attention mismatch after compression caused by token reduction.
To verify this mechanism, we analyze all 32 LLM layers on the full VQA$^{\text{T}}$ validation set with LLaVA-1.5-7B under the $576 \rightarrow 144$ compression setting.
We compare the unpruned model, the compressed model without LAR, and the compressed model with LAR.
The compressed model without LAR is denoted as Collapsed.
We measure relative attention logit deviation and token-group KL divergence against the unpruned reference.
As shown in Fig.~\ref{fig:extra_lar_mechanism}, Collapsed systematically underestimates visual attention after token merging.
After applying LAR, the mean absolute logit deviation decreases from 23.0\% to 8.2\%, and several layers move close to the dense-reference region.
At the group-distribution level, the mean token-group KL decreases from 0.90 to 0.40. The layer-wise recovery reaches 68\% for logit error and 54\% for group-level KL.
These results show that LAR directly restores attention statistics distorted by token reduction, rather than only improving the final accuracy.
\subsection{Efficiency and Deployment Analysis}
\label{ssec:efficiency}
In this section, we examine the efficiency and deployment analysis of our proposed method, covering computational cost, LAR overhead and vLLM~\cite{kwon2023efficient} deployment benefits.
\\
\noindent\textbf{Analysis of Efficiency.}
Tab.~\ref{tab:efficiency_comparison_1_6_final} compares computational efficiency and retained accuracy.
With ERA, reducing 2{,}880 visual tokens to 160 lowers FLOPs by $11.8\times$, from 66.1 TFLOPs to 5.6 TFLOPs. 
The end-to-end (E2E) latency improves by $2.7\times$, and prefill is accelerated by $4.6\times$, decreasing from 415 ms to 91 ms.
The KV cache memory drops from 1.44 GB to 0.08 GB.
Beyond efficiency, ERA achieves 95.8\% performance, which is 5.3\% higher than DART.
These results show that ERA provides a better trade-off.
\\
\noindent\textbf{LAR Overhead in Standard Inference.}
We further verify the practical overhead of our LAR on GQA, using a compressed visual sequence of 160 tokens with FlashAttention v2.7.3.
Tab.~\ref{tab:extra_lar_overhead} shows that LAR introduces no measurable overhead in end-to-end inference. 
The latency difference remains within measurement noise while the peak GPU memory usage is unchanged.
Meanwhile, the performance on GQA improves by 0.87\%.
These results confirm that LAR can be deployed without incurring additional latency or memory cost, while still providing a significant accuracy boost.
%
% Although isolated attention microbenchmarks for a single layer can show visible relative overhead due to very small kernel times, the absolute cost is diluted in full inference by MLP computation, vision encoding, projection, layer normalization and runtime scheduling.
\\
\noindent\textbf{vLLM Deployment Results.}
We further evaluate ERA in a practical vLLM serving setup using LLaVA-NeXT-7B.
As shown in Tab.~\ref{tab:extra_vllm_deployment_rowwise}, ERA consistently improves deployment efficiency across different batch sizes while remaining compatible with the standard vLLM serving pipeline.
At Batch=256, ERA reduces the prefill sequence length from 1845 to 373 tokens, leading to a $4.3\times$ reduction in LLM prefill latency and a $5.0\times$ reduction in per-request KV cache usage.
This memory reduction is particularly important for high-throughput serving. Under a full batching, the original baseline requires approximately 244 GB of GPU memory, which exceeds the capacity of a single 80 GB A800 GPU. In contrast, ERA reduces the peak memory footprint to only 59.69 GB, enabling the entire batch to be served on a single device without batch partitioning.
The deployment benefit also generalizes across different scheduling regimes. ERA achieves $3.2\times$ prefill acceleration at Batch=48 and still maintains a $2.4\times$ speedup at Batch=1.
Importantly, the decode throughput remains nearly unchanged, indicating that ERA primarily accelerates the prefill stage without introducing additional overhead to the decode path.
In the same vLLM deployment setting, ERA also preserves task performance under aggressive token pruning, improving VQA$^{\text{T}}$ accuracy from 49.19\% to 56.87\% and POPE average accuracy from 86.59\% to 88.25\%.
Overall, these results demonstrate that ERA provides practical deployment advantages while maintaining compatibility with optimized vLLM pipelines.
\subsection{Sensitivity Analysis}
\label{ssec:hyper}
In this section, we examine the sensitivity of ERA to its main hyperparameters: the balance factor $\alpha$ and the regularization weight $\eta$.
Meanwhile, we also explore the effect of pruning depths on the encoder-free architecture NEO.
\begin{figure}[t]
  \centering
  \vspace{-2mm}
  \includegraphics[width=1\linewidth]{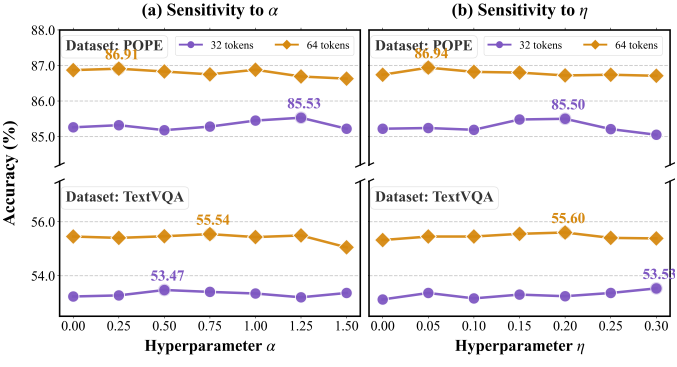}
  \caption{Hyperparameter sensitivity of ERA with respect to $\alpha$ and $\eta$.}
  \label{fig:hyperparameters}
\end{figure}
\begin{figure}[!t]
    \centering
    \vspace{-3mm}
    \includegraphics[width=0.485\textwidth]{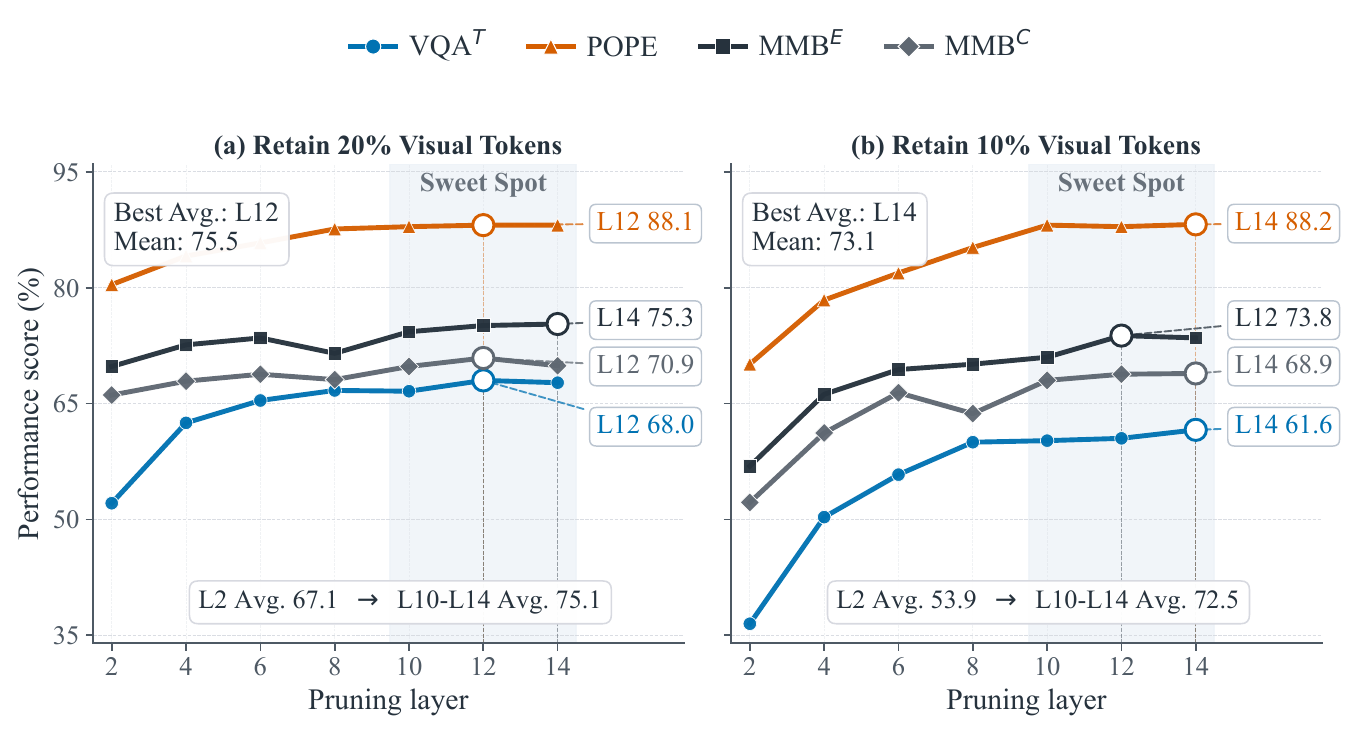}
    \caption{Pruning-depth sensitivity with \textbf{NEO-2B}.}
    \label{fig:neo_ablation}
\end{figure}
\begin{figure*}[!t]
    \centering
    \includegraphics[width=0.9\textwidth]{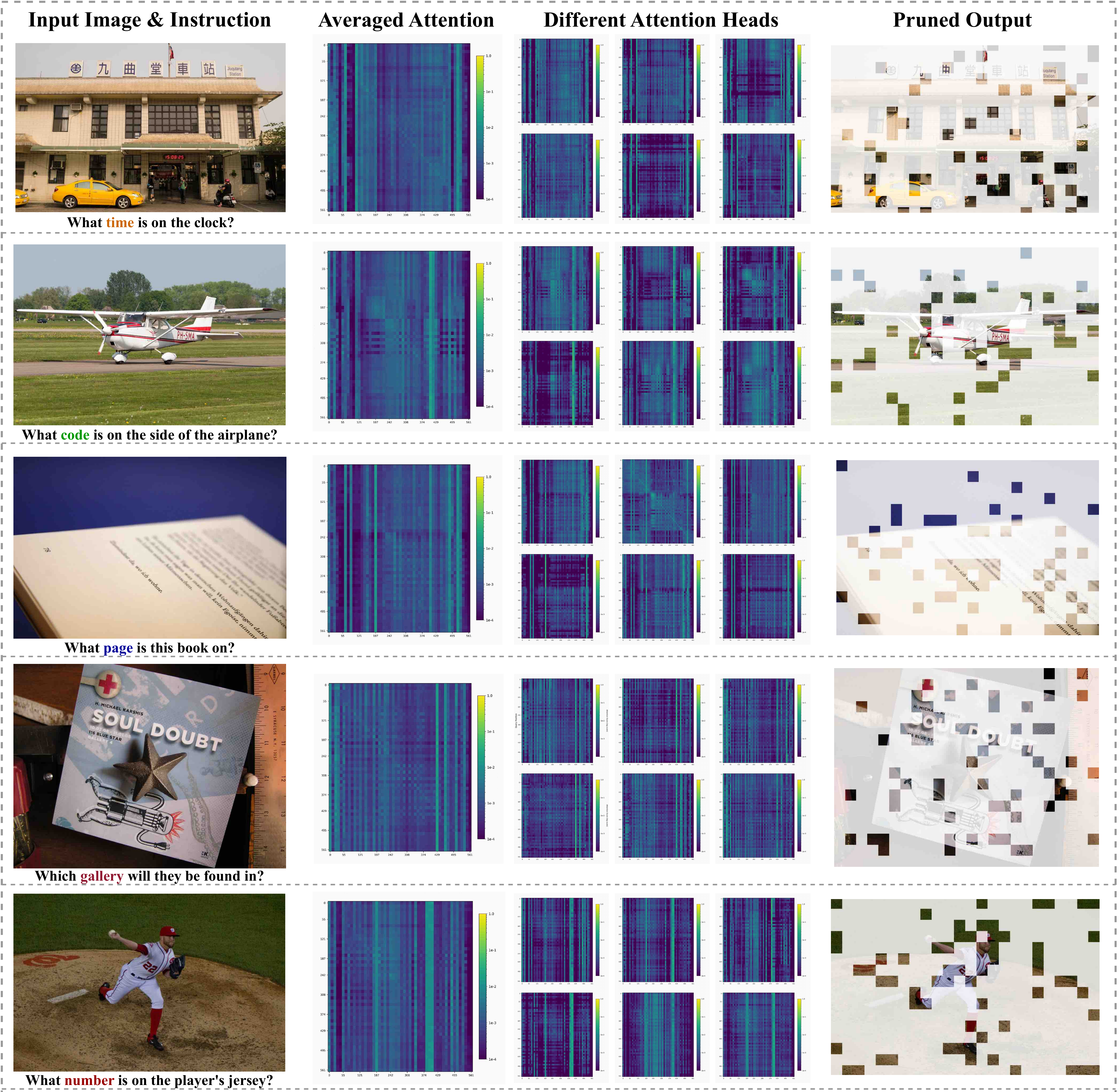}
    \caption{Head-wise attention visualizations and token pruning results with \textbf{LLaVA-1.5-7B} on the VQA$^{\text{T}}$ dataset.}
    \label{fig:appendix_fig1}
    \vspace{-2mm}
\end{figure*}
\begin{figure}[t]
  \centering
  \includegraphics[width=1\linewidth]{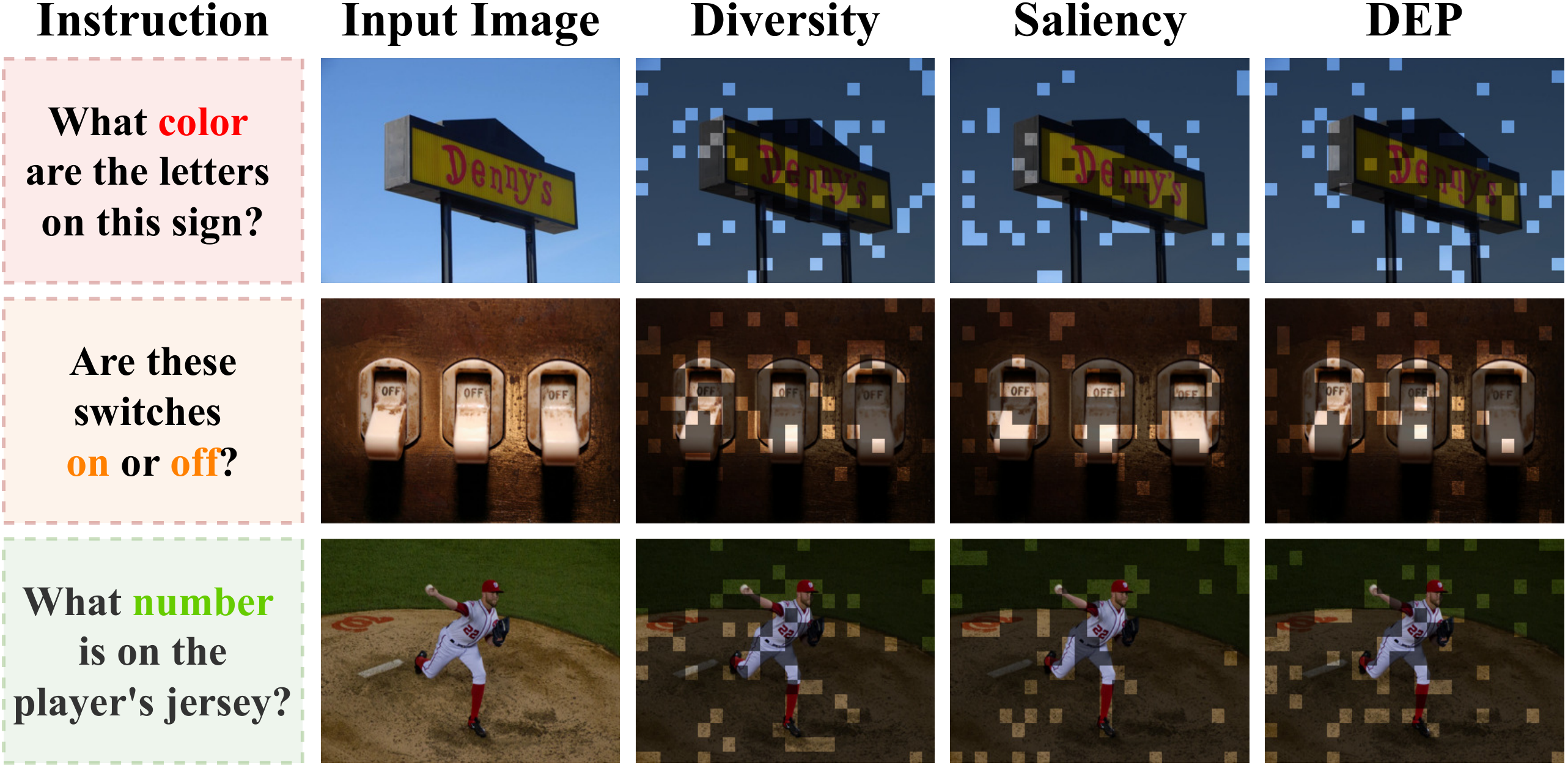}
  \caption{Visualization of token pruning under different DEP criteria.}
  \label{fig:vis_token_pruning}
  \vspace{-1mm}
\end{figure}
\\
\noindent\textbf{Analysis of Hyperparameters.}
Fig.~\ref{fig:hyperparameters} analyzes the sensitivity of ERA to $\alpha$ and $\eta$.
Overall, ERA remains stable across a wide range of values for both hyperparameters.
Specifically, Fig.~\ref{fig:hyperparameters} (a) indicates that varying $\alpha$ yields consistently strong performance across datasets.
In both the 32-token and 64-token settings, the curves fluctuate only mildly across the tested values.
Fig.~\ref{fig:hyperparameters} (b) shows that the performance is insensitive to $\eta$.
Across datasets and token budgets, the curves remain stable, indicating that the entropy-guided regularization is not sensitive to the hyperparameter values.
\\
\noindent\textbf{Pruning-depth Verification in NEO.}
Unlike modular MLLMs that decouple vision encoders from language backbones, NEO integrates raw visual patches directly into a unified transformer with naive attention.
In this setting, the pruning depth determines when the visual-token sequence is compressed.
Fig.~\ref{fig:neo_ablation} summarizes an ablation over pruning depths on NEO-2B.
Pruning too early degrades performance across benchmarks. At 20\% token retention, the average score rises from 67.1\% at layer 2 to 75.1\% across layers 10 to 14. At 10\% retention, the corresponding average rises from 53.9\% to 72.5\%.
In contrast, pruning at deeper layers yields stable results, with the stronger-performance region spanning layers 10 to 14 around the end of the pre-Buffer stage.
This phenomenon matches the design of NEO, where the pre-Buffer stage ends at the 12th layer.
Thus, we choose the 12th layer as the default pruning depth for different pruning methods on NEO-2B for fair comparison.
\subsection{Qualitative Visualizations}
\label{ssec:vis}
In the final section of experiments, we provide qualitative visualizations of ERA's head-wise attention distribution and token pruning results under different DEP criteria in Tab.~\ref{tab:ablation_main}.
\\
\noindent\textbf{Bias of Head-averaged Attention.}
Fig.~\ref{fig:appendix_fig1} presents a qualitative comparison between averaged attention maps and the activation patterns of individual attention heads across various inputs.
Observing the \textit{Averaged Attention} column, the heatmaps exhibit a smoothed distribution that fails to distinctively highlight specific regions.
In contrast, the visualization of \textit{Different Attention Heads} reveals that semantic information is sparsely encoded in specific heads. 
These heads act as sharp detectors for instruction-aligned features, such as the clock face or airplane code, but their signals are diluted when aggregated.
This discrepancy shows that simple head-averaging can mask critical tokens.
As shown in the \textit{Pruned Output}, DEP mitigates this averaging bias. 
It isolates the visual anchors required to answer the prompt while discarding background redundancy.
These results support the motivation for incorporating head-wise attention entropy as a novel criterion for token importance estimation, generating more reliable pruning decisions.
\\
\noindent\textbf{Visualization of Token Pruning.}
Fig.~\ref{fig:vis_token_pruning} visualizes token pruning results based on different DEP criteria.
Diversity-based DEP focuses on high-variance background regions, whereas saliency-based DEP favors the dominant object but may miss fine-grained details.
DEP jointly balances saliency and diversity, leading to more reliable token pruning.
\section{Conclusion}
\label{sec:conclusion}
In this paper, we revisit visual token reduction in MLLMs from the perspective of attention behavior.
We show that effective token reduction requires not only selecting a compact set of visual tokens, but also preserving a compatible attention distribution after compression to avoid the overlooked phenomenon we term \emph{Attention Logit Collapse}.
To this end, we propose \textbf{ERA}, a training-free token reduction framework that integrates entropy-guided token pruning with rectified attention.
Specifically, we leverage \textbf{Dual-view Entropy Pruning (DEP)} to capture both diversity and saliency.
Then, we employ \textbf{Bias-aware Token Recycling (BTR)} to recycle tokens by estimating a bias in each cluster.
Finally, we adopt \textbf{Logit-preserving Attention Rectification (LAR)} to maintain the distribution in a kernel-friendly manner.
Extensive experiments across MLLM architectures, resolutions and benchmark settings demonstrate that ERA achieves robust and efficient inference under aggressive token reduction.
Beyond accelerating MLLM inference, ERA highlights \emph{Attention Logit Collapse} as a neglected failure mode in current training-free visual token reduction methods and provides a unified solution across theory, algorithm and deployment.

\bibliographystyle{IEEEtran}
\bibliography{refs}
\begin{IEEEbiography}[{\includegraphics[width=1in,height=1.24in,clip,keepaspectratio]{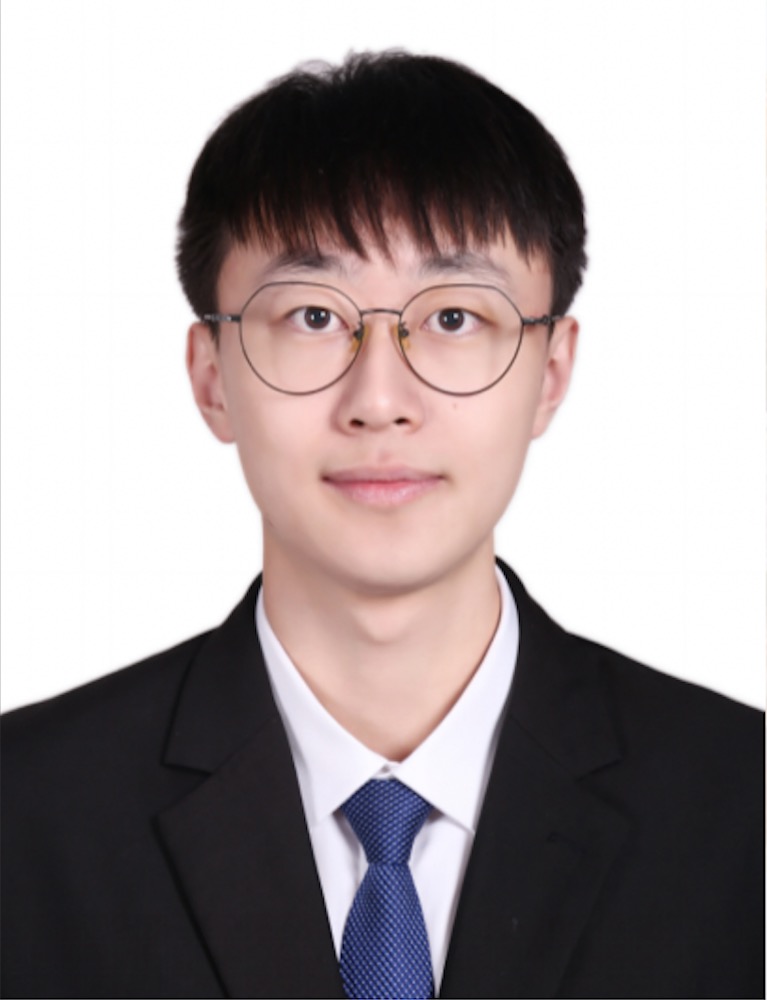}}]{Yuhao Wang} received the B.E. degree in Artificial Intelligence from the School of Future Technology, Dalian University of Technology (DUT), Dalian, China, in 2024. He is pursuing the M.S. degree in Artificial Intelligence at the School of Computer Science, DUT. His research interests include efficient large language models and multimodal fusion.
\end{IEEEbiography}
\begin{IEEEbiography}[{\includegraphics[width=1in,height=1.24in,clip,keepaspectratio]{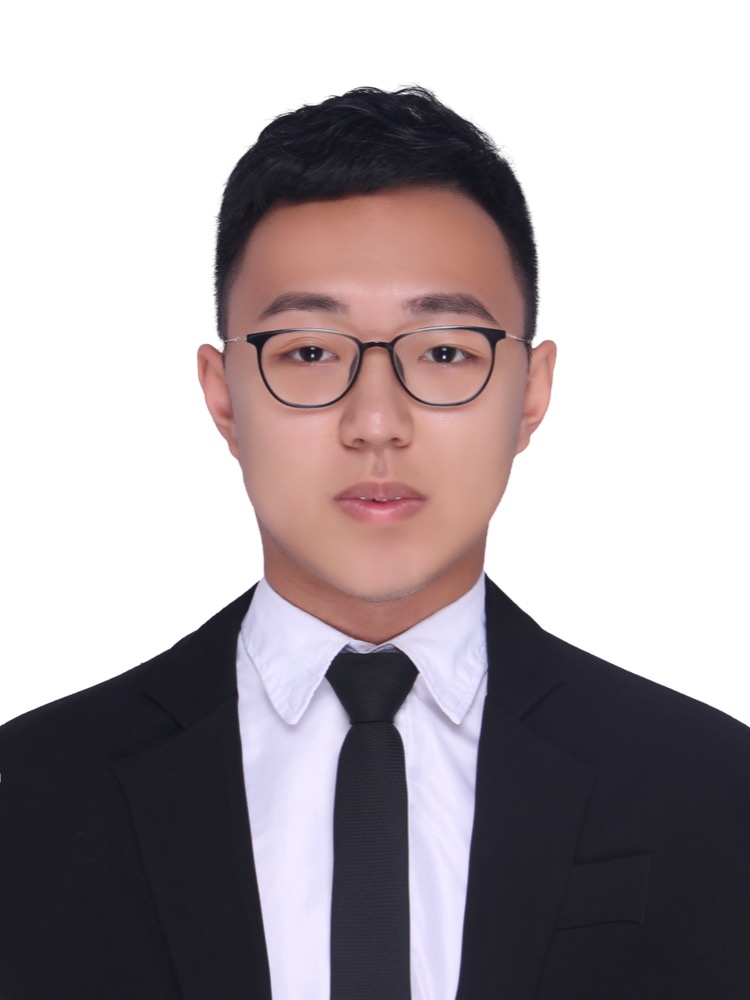}}]{Mu Qiao}
    is currently pursuing the B.E. degree in Software Engineering at the School of Software, Dalian University of Technology (DUT), Dalian, China. His research interests include multimodal large models and efficient vision-language models.
    \end{IEEEbiography}
\begin{IEEEbiography}[{\includegraphics[width=1in,height=1.24in,clip,keepaspectratio]{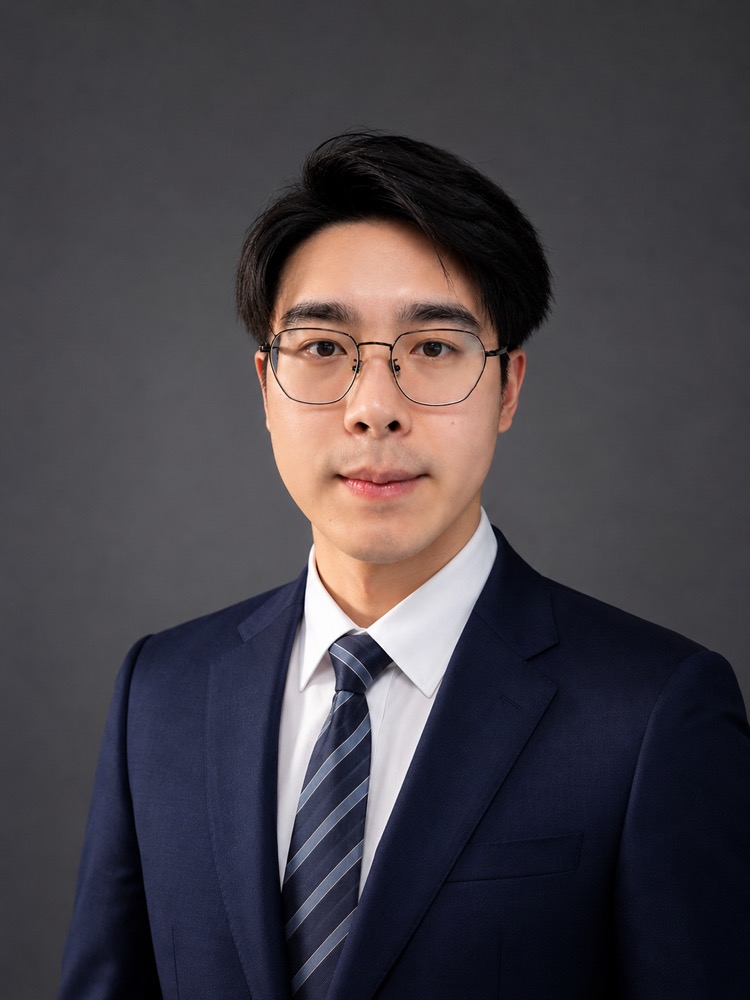}}]{Haiwen Diao} received the B.S., M.S., and Ph.D. degrees from Dalian University of Technology (DUT), Dalian, China. He is currently a Research Fellow at MMLab, Nanyang Technological University (NTU), Singapore. His research interests include computer vision and artificial intelligence, with a focus on native multimodal models, vision-language understanding and generation, and inference-time reasoning.
\end{IEEEbiography}
\begin{IEEEbiography}[{\includegraphics[width=1in,height=1.24in,clip,keepaspectratio]{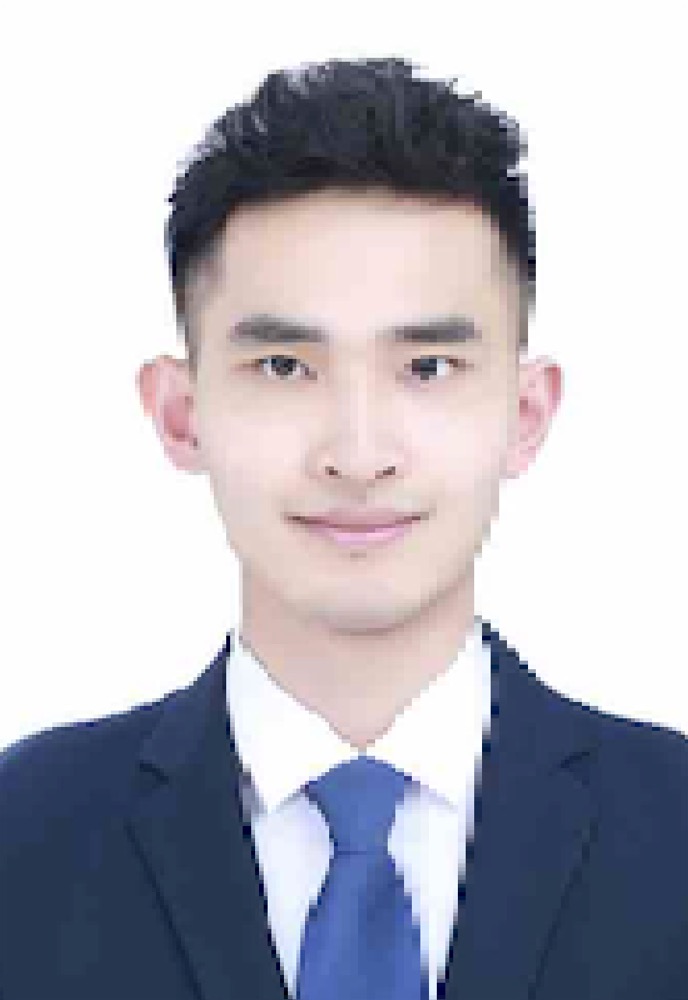}}]{Yunzhi Zhuge} received the B.E. degree in electrical engineering from Shandong University of Science and Technology, Qingdao, China, in 2017, the M.S. degree in biomedical engineering from Dalian University of Technology, Dalian, China, in 2019, and the Ph.D. degree in computer science from the University of Adelaide, Adelaide, SA, Australia, in 2023. He is currently a postdoctoral researcher at Dalian University of Technology.
\end{IEEEbiography}
\begin{IEEEbiography}[{\includegraphics[width=1in,height=1.24in,clip,keepaspectratio]{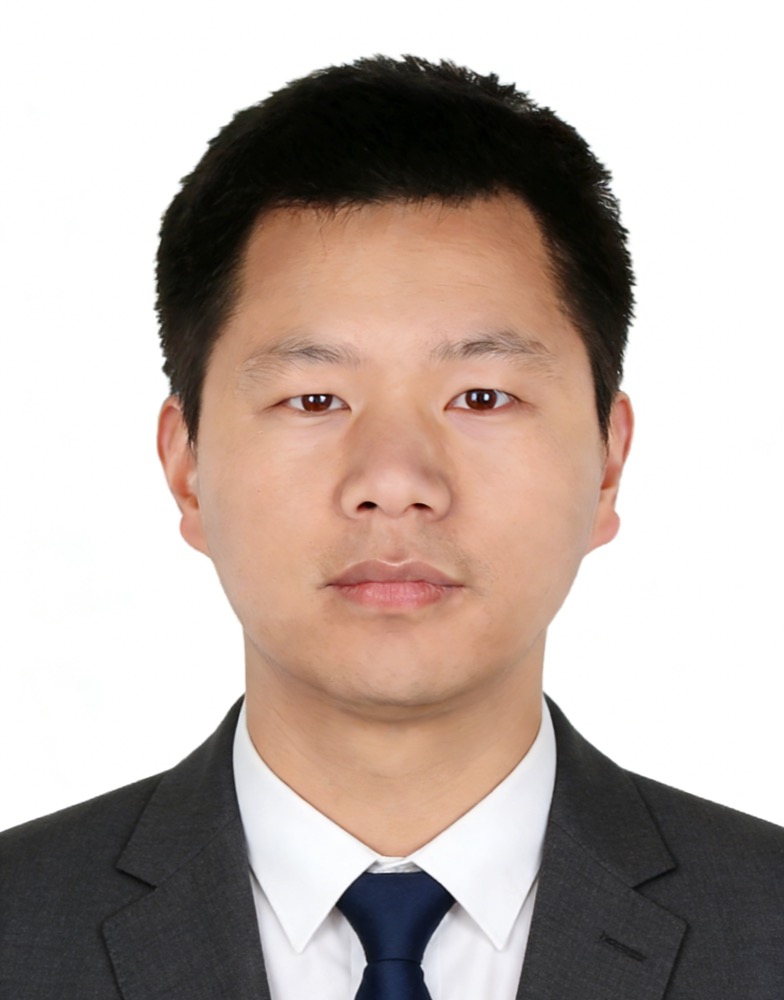}}]{Pingping Zhang} received the B.E. degree in mathematics and applied mathematics from Henan Normal University (HNU), Xinxiang, China, in 2012, and the Ph.D. degree in signal and information processing from the Dalian University of Technology (DUT), Dalian, China, in 2020. He is currently an Associate Professor with the School of Future Technology and the School of Artificial Intelligence, DUT. He has authored over 70 papers in top-tier journals and conferences. His research interests include deep learning, saliency detection, person re-identification, and semantic segmentation.
\end{IEEEbiography}
\begin{IEEEbiography}[{\includegraphics[width=1in,height=1.24in,clip,keepaspectratio]{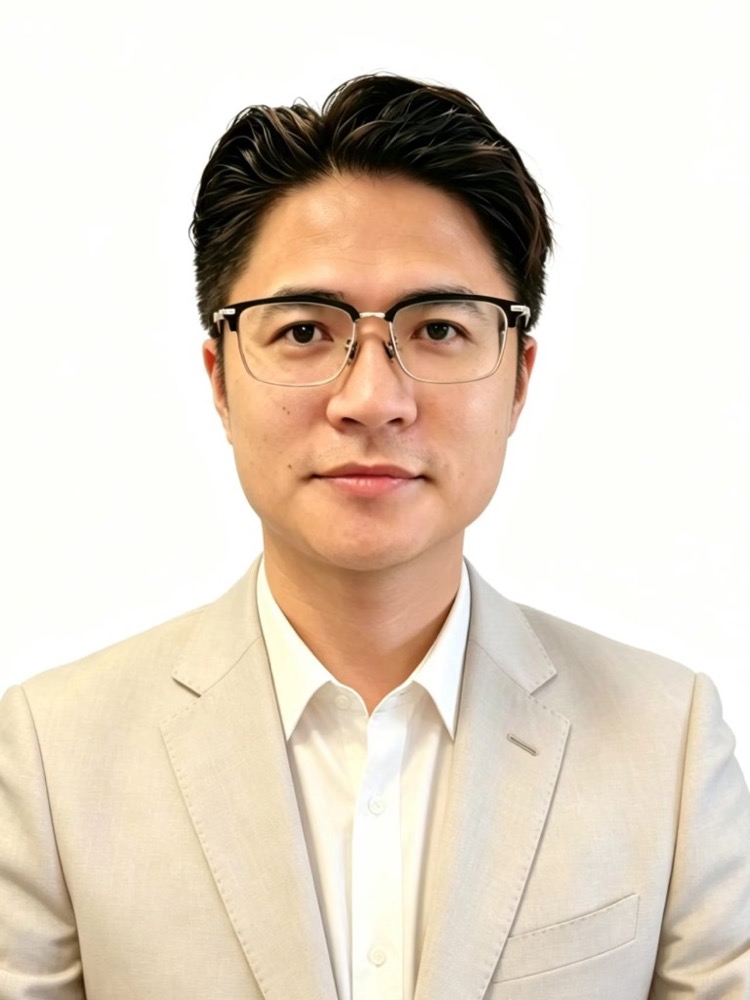}}]{Xindong Zhang} received the B.S., M.S., and Ph.D. degrees from Xiamen University, Xiamen, China, Dalian University of Technology (DUT), Dalian, China, and The Hong Kong Polytechnic University, Hong Kong, respectively. He is currently with OPPO, where he works on AI ISP algorithm engineering and technical planning. His research interests include numerical optimization and applications, lightweight AI model design, and AI-based computational imaging.
\end{IEEEbiography}
\begin{IEEEbiography}[{\includegraphics[width=1in,height=1.24in,clip,keepaspectratio]{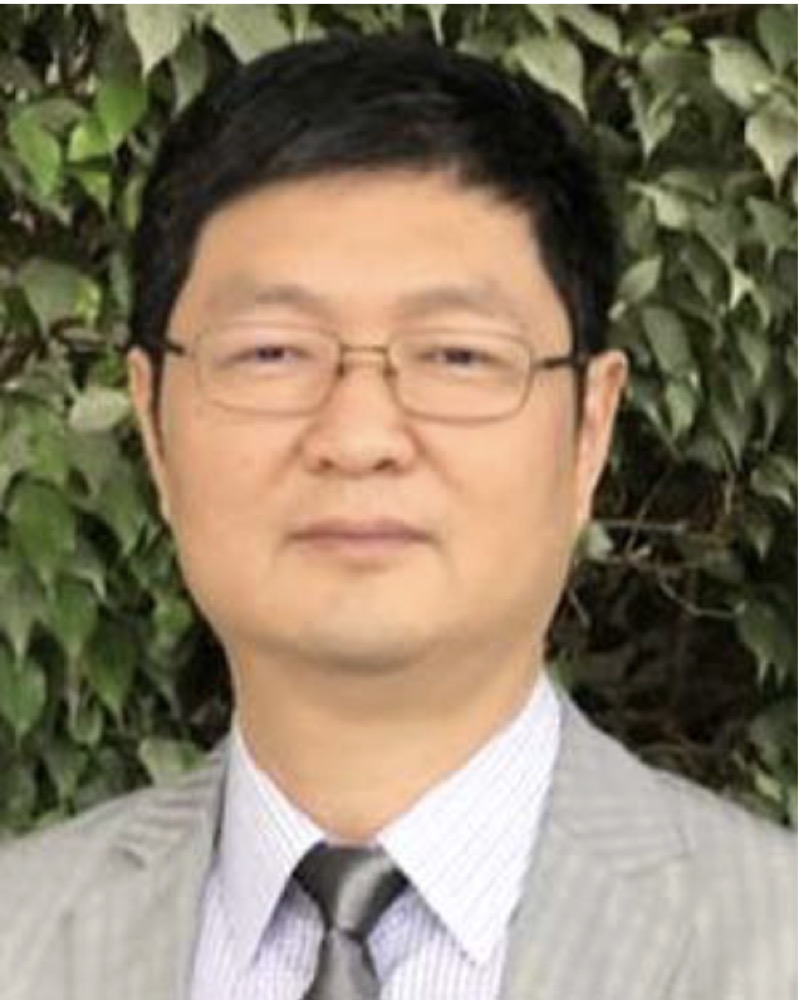}}]{Lei Zhang} (IEEE Fellow) received the B.Sc. degree from Shenyang Institute of Aeronautical Engineering, Shenyang, China, in 1995, and the M.Sc. and Ph.D. degrees in control theory and engineering from Northwestern Polytechnical University, Xi'an, China, in 1998 and 2001, respectively. From 2001 to 2002, he was a Research Associate with the Department of Computing, The Hong Kong Polytechnic University. From 2003 to 2006, he was a Postdoctoral Fellow with the Department of Electrical and Computer Engineering, McMaster University, Canada. In 2006, he joined the Department of Computing, The Hong Kong Polytechnic University, as an Assistant Professor. Since 2017, he has been a Chair Professor in the same department. His research interests include computer vision, image and video analysis, pattern recognition, and biometrics. He has published more than 200 papers in these areas. As of 2022, his publications have been cited more than 80,000 times. He is a Senior Associate Editor of IEEE Transactions on Image Processing and has served as an Associate Editor of IEEE Transactions on Pattern Analysis and Machine Intelligence, SIAM Journal on Imaging Sciences, IEEE Transactions on Circuits and Systems for Video Technology, and Image and Vision Computing. He was named a Clarivate Analytics Highly Cited Researcher from 2015 to 2022.
\end{IEEEbiography}
\begin{IEEEbiography}[{\includegraphics[width=1in,height=1.24in,clip,keepaspectratio]{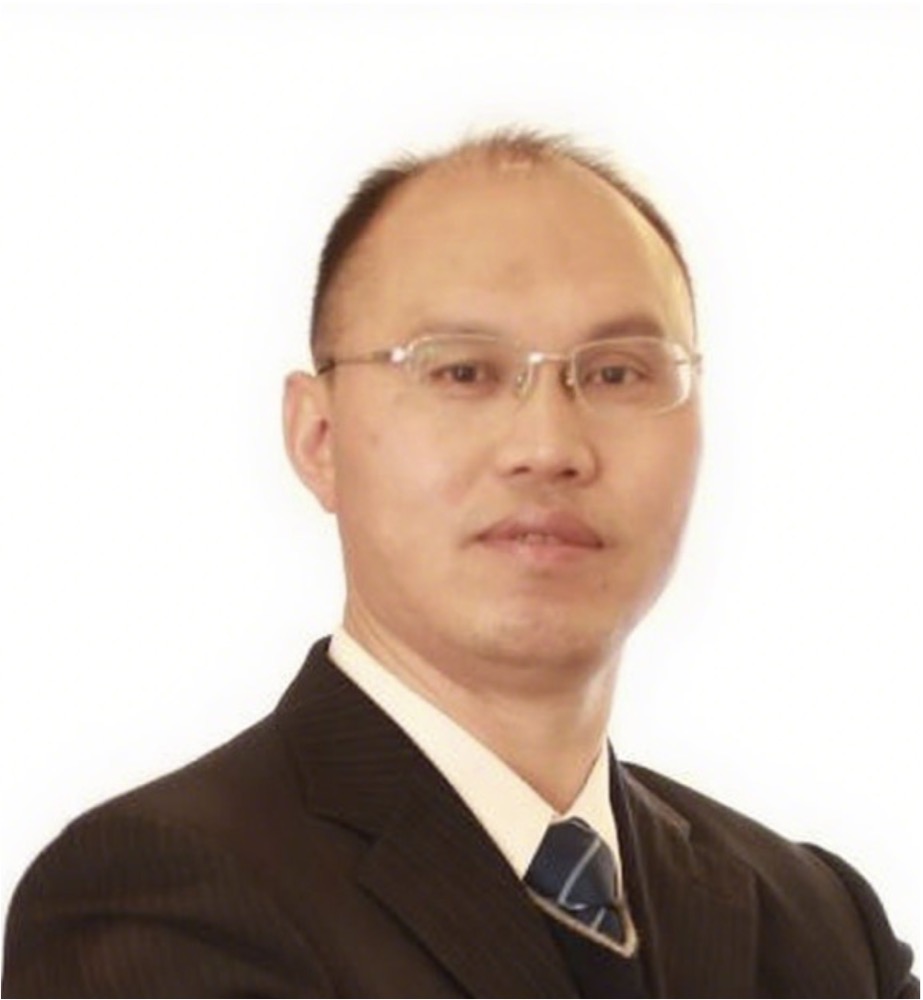}}]{Huchuan Lu} (IEEE Fellow) received the MS degree in signal and information processing and the Ph.D. degree in systems engineering from the Dalian University of Technology (DUT), Dalian, China, in 1998 and 2008, respectively. In 1988, he joined the faculty of the School of Information and Communication Engineering, DUT, where he is currently a full professor. He is also the Dean of the School of Future Technology, DUT. His current research interests include computer vision and artificial intelligence with a focus on visual tracking, and multimodal large language models.
\end{IEEEbiography}
\end{document}